\documentclass[11pt]{article}

\PassOptionsToPackage{hyperfootnotes=false}{hyperref}
\usepackage[final]{acl}

\usepackage{times}
\usepackage{latexsym}
\usepackage[T1]{fontenc}
\usepackage[utf8]{inputenc}
\usepackage{microtype}
\usepackage{inconsolata}
\usepackage{graphicx}
\usepackage{placeins}
\usepackage{needspace}
\usepackage{booktabs}
\usepackage{amsmath}
\usepackage{amssymb}
\usepackage{xcolor}
\usepackage{algorithm}
\usepackage{algorithmic}
\usepackage{multirow}
\usepackage{enumitem}
\usepackage{tcolorbox}
\usepackage{xspace}
\usepackage[toc,page,header]{appendix}
\usepackage{minitoc}
\usepackage{adjustbox}
\usepackage{setspace}
\usepackage{subcaption}
\usepackage{tabularx}
\usepackage{bm}
\doparttoc
\faketableofcontents
\tcbuselibrary{skins,breakable}

\newtcolorbox{examplebox}[1]{
  colback=gray!5, colframe=gray!60, title=#1,
  fonttitle=\bfseries\small, sharp corners, boxrule=0.5pt
}
\newtcolorbox{instrbox}{
  colback=blue!5, colframe=blue!40, title=Instruction,
  fonttitle=\bfseries\scriptsize, sharp corners, boxrule=0.4pt, left=4pt, right=4pt, top=2pt, bottom=2pt
}
\newtcolorbox{pickbox}[1]{
  colback=red!5, colframe=red!40, title=#1,
  fonttitle=\bfseries\scriptsize, sharp corners, boxrule=0.4pt, left=4pt, right=4pt, top=2pt, bottom=2pt
}
\newtcolorbox{goldbox}[1]{
  colback=green!5, colframe=green!40, title=#1,
  fonttitle=\bfseries\scriptsize, sharp corners, boxrule=0.4pt, left=4pt, right=4pt, top=2pt, bottom=2pt
}

\newcommand{\tool}{\mbox{\textsc{REE-BoN}}\xspace}

\title{\textsc{REE-BoN}: Efficient Best-of-N Scoring via Depth-wise Retrieve-Rerank}

\author{%
  Tianyi Li\textsuperscript{1} \quad
  Aditya A. Ramesh\textsuperscript{2}\thanks{\ Work done while at Amazon.} \quad
  Nacime Bouziani\textsuperscript{3}\footnotemark[1] \\
  \bfseries Jens Lehmann\textsuperscript{1} \quad
  Stefanos Laskaridis\textsuperscript{1} \\
  \normalfont \textsuperscript{1}Amazon Science \\
  \normalfont \textsuperscript{2}The Swiss AI Lab, IDSIA USI-SUPSI \quad
  \textsuperscript{3}HUMAIN, Saudi Arabia \\
  \normalfont\small \texttt{tylteddy@amazon.com}, 
  \texttt{aditya@idsia.ch}, \texttt{nbouziani@humain.com},
  \texttt{jlehmnn@amazon.com}, 
  \texttt{mail@stefanos.cc}
}

\begin{document}

\maketitle

\begin{abstract}
Best-of-N selection scores $N$ candidates with a reward model, requiring $N$ full forward passes. Since intermediate transformer layers are known to carry rich task-relevant information, we ask whether early layers of a reward model can filter candidates before committing to full-depth evaluation. We propose \tool, a ``retrieve-rerank'' pipeline along the depth axis of a reward model: shallow layers cheaply narrow the candidate set, and deeper layers discriminate among the shortlist. To make early layers reliable filters, we train lightweight exit heads on a frozen backbone with a margin-weighted multi-response Bradley-Terry loss. We jointly optimize which layers filter and how many candidates each keeps on a small calibration set, but also provide a zero-calibration alternative. With a default setting shared by all 20 configurations, spanning five reward models, two preference datasets and two policies, the pipeline cuts reward-model scoring FLOPs by 41\% on average. In 17 of the 20 configurations, its selection quality is at most 1\% below full depth, on a scale normalized between random and oracle selection. When savings are reinvested into additional candidates, the pipeline improves selections at matched budget.
\end{abstract}

\section{Introduction}
\label{sec:intro}

\begin{figure}[t]
\centering
\includegraphics[width=\columnwidth,trim=0.35cm 0.75cm 0.35cm 0.2cm, clip]{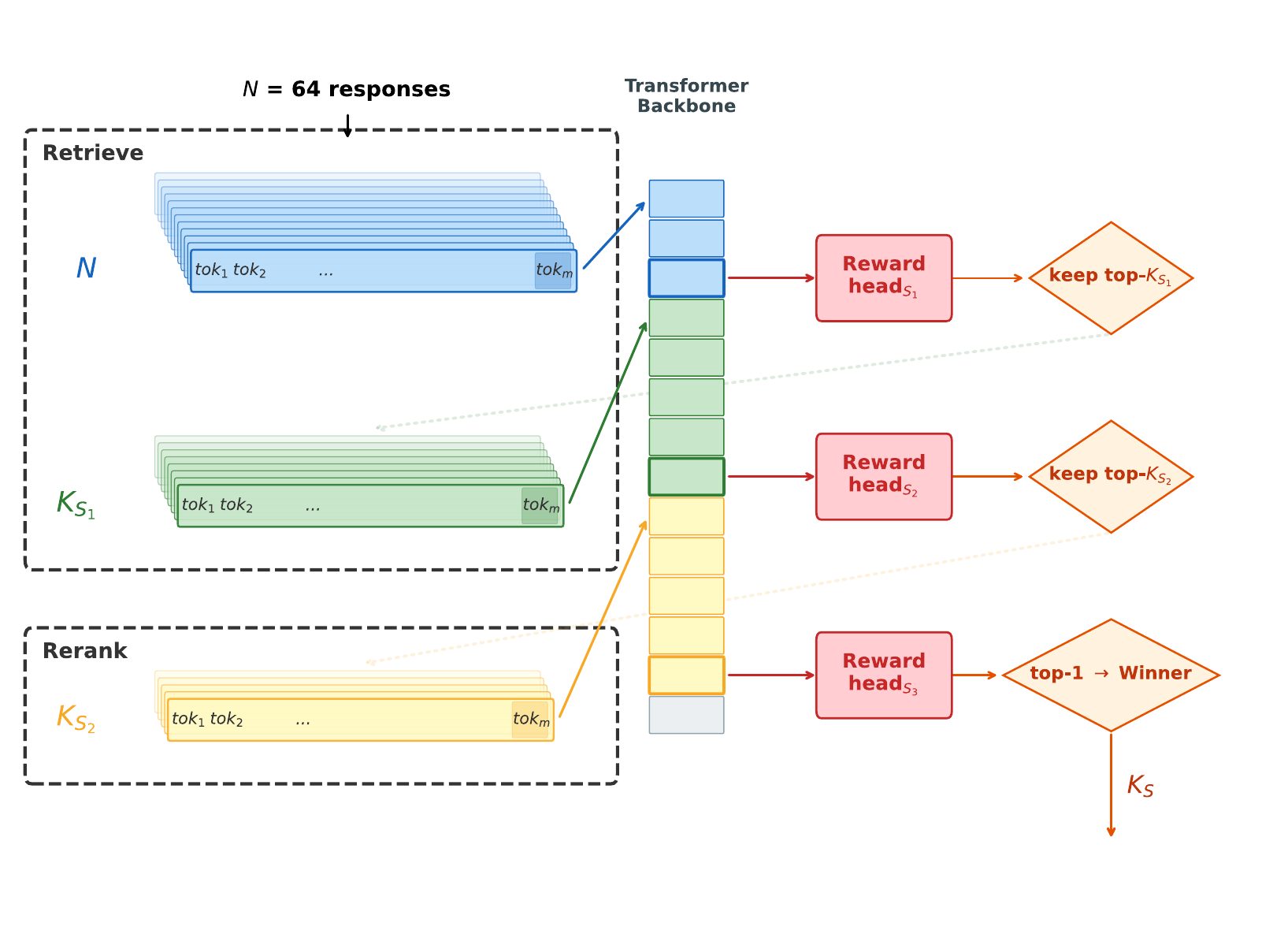}
\vspace{-0.2in}
\caption{Overview of the \tool pipeline. $N$ = 64 response sequences are processed through transformer layers (dashed box). At selected layers, a lightweight reward head scores all surviving candidates and a filter retains the top-$K_s$. Only survivors continue to deeper layers, progressively narrowing the candidate set until a single winner is selected. No layers are recomputed.}
\label{fig:pipeline-overview}
\vspace{-0.1in}
\end{figure}

Best-of-N (BoN) selection is a simple and widely adopted strategy for inference-time alignment: generate $N$ responses from a base policy, score each with a reward model, and return the highest-scoring candidate. The same pattern arises in, e.g., rejection sampling pipelines~\citep{Touvron2023Llama2,dongraft}, where many candidates are scored and filtered to curate training data. In both cases, scoring cost scales linearly with the number of candidates, requiring $N$ full forward passes through the reward model, which can be costly and potentially redundant, especially at the edge~\cite{10591422}.

In large language models (LLMs), deeper layers exhibit striking redundancy. \citet{Gromov2025Layers} show that contiguous blocks of later layers can be pruned from generative LLMs with minimal QA degradation, and \citet{Belrose2023TunedLens} use probes to track how token predictions change across layers.
Early-exit methods exploit this redundancy for generation \citep{Schuster2022CALM,Elhoushi2024LayerSkip} and classification \citep{Fan2024AdaInfer}. Best-of-N selection can use this redundancy by asking shallow layers only to filter, an easier task than selection, so that only a shortlist needs deeper evaluation. We exploit this asymmetry and develop the training recipe and inference strategy needed to make intermediate layer scoring effective for BoN \mbox{selection.}

On the training side, we start from a trained reward model (RM), freeze its backbone, and train lightweight MLP exit heads at every intermediate layer with a margin-weighted multi-response Bradley-Terry loss over all $K$ responses per prompt. This two-stage recipe gives better early and middle layers than training the backbone and heads together (\S\ref{sec:freeze}). The multi-response loss lifts early-layer quality by ${\sim}4\%$ of the oracle$-$random reward range compared to pairwise alternatives.

On the inference side, we propose a ``retrieve-rerank'' cascade along layer depth. Shallow layers cheaply filter the candidate set, and deeper layers discriminate among a progressively smaller shortlist. The pipeline has two degrees of freedom: \textit{(i)}~which layers serve as stages, and \textit{(ii)}~how many candidates each stage retains. We optimize these jointly over a small calibration set, and also provide a zero-calibration heuristic as a practical default. Because all exit heads share one common weight-shared backbone, the entire pipeline runs within a single deployed model, without the need for multi-model routing or additional serving infrastructure overhead.

We propose \tool (Reward modeling with Early Exiting for BoN, pronounced ``ribbon''), which has two components: \textit{(1)}~frozen-backbone training of exit heads with a multi-response comparison loss; and \textit{(2)}~a calibrated depth-wise ``retrieve-rerank'' pipeline that jointly selects exit layers and retention rates. Our experiments support three findings. First, \tool keeps full-depth quality at a much lower cost: with one default setting ($\lambda{=}0.05$) for all configurations, it cuts RM scoring FLOPs by 41\% on average, and in 17 of the 20 configurations its selection quality is at most 1\% below full depth (\S\ref{sec:pareto}). Second, it is practical to deploy: a zero-calibration heuristic matches these savings at 0.5\% mean degradation, only 0.2\% above the calibrated pipeline (\S\ref{sec:pareto}), and calibrated pipelines are robust to dataset and policy shift (\S\ref{sec:shift}). Third, reinvesting the savings in more candidates improves selections at matched budget (\S\ref{sec:budget}).

\section{Related Work}
\label{sec:related}

\paragraph{Early exiting and adaptive depth.}
\citet{laskaridis2021adaptive} survey early exits in deep neural networks. At the \emph{layer level}, AdaInfer \citep{Fan2024AdaInfer} adaptively exits at an intermediate layer per input on classification tasks, and LayerSkip \citep{Elhoushi2024LayerSkip} trains with graduated layer dropout to enable early exit and self-speculative decoding. At the \emph{token level}, CALM \citep{Schuster2022CALM} exits per-token based on confidence, SkipDecode \citep{DelCorro2023SkipDecode} provides batch-compatible exit schedules, and Mixture of Depths \citep{Raposo2024MoD} routes tokens to a subset of layers via learned top-$k$ selection. On encoders, confidence-based exits \citep{xin-etal-2020-deebert} and PABEE's patience rule \citep{zhou2020bert} stop computation per input once intermediate classifiers are confident or agree. These methods decide the exit depth of each input or token on its own. We instead score the candidates for the same prompt together and pass only the top-ranked ones to deeper layers.

\paragraph{Cascades for ranking.}
The Cascade Transformer \citep{soldaini-moschitti-2020-cascade} follows the same paradigm as our pipeline: to select answers among retrieved sentences, it builds a retrieve-rerank cascade in one transformer, attaching classifiers to intermediate layers and dropping low-scoring candidates as the batch moves deeper. For reward models in Best-of-N selection, we show how to train exit heads that filter reliably and how to calibrate the cascade for a chosen cost--quality trade-off.

\paragraph{Reward model efficiency.}
Several approaches aim to reduce reward scoring cost. TinyRM \citep{Pan2025TinyRM} trains small specialist RMs that rival larger models on specific domains. \citet{Xu2025AskStrongJudge} route between a fast RM and a costly LLM judge based on uncertainty. Fast-Slow Thinking RM \citep{Wu2026FastSlowRM} integrates scalar and chain-of-thought scoring modes in one model. These methods use separate models, separate modes, or autoregressive reasoning. In contrast, we leverage depth hierarchy within a single scalar-output model, so no additional model or generation is needed.

\paragraph{Multi-head and ensemble reward models.}
\mbox{ArmoRM} \citep{Wang2024ArmoRM} decomposes reward into 19 interpretable objectives via separate heads on a frozen backbone, combining them with learned gating. \citet{Eisenstein2024RMEnsembles} show that RM ensembles mitigate reward hacking, though at the cost of maintaining multiple full models; WARM \citep{Rame2024WARM} addresses this by averaging RM weights into a single model. Our exit heads also share a frozen backbone, but decompose by \emph{depth} rather than by objective, yielding a hierarchy of increasingly refined scores from a single forward pass. \citet{leontiadis2021s} personalize exit heads in CNNs. Whether this helps reward models remains open.

\paragraph{Best-of-N efficiency.}
AdaBoN \citep{Raman2025AdaBoN} adaptively allocates generation budget across prompts based on estimated difficulty. \mbox{TreeBoN} \citep{TreeBoN2024} uses iterative branch-and-prune to improve candidate quality at fixed budget. Speculative Rejection \citep{Sun2024SpeculativeRejection} terminates generation of low-scoring candidates early. These approaches reduce the cost of producing candidates; our work reduces the cost of scoring them. The two axes are orthogonal and composable.

\paragraph{Multi-response and margin-weighted preference learning.}
Training on all $K$ responses per prompt to obtain $\binom{K}{2}$ preference pairs was used by \citet{Ouyang2022InstructGPT}. Using the preference margin in the pairwise loss was explored by \citet{Touvron2023Llama2} and \citet{Wang2025HelpSteer2Pref}; the approach is analogous to importance weighting in learning-to-rank \citep{Burges2006LambdaRank}. 
We find that margin-weighted multi-response training is critical specifically for early-exit heads, where it enables shallow layers to extract ranking signal that standard pairwise losses miss.

\section{Method}
\label{sec:method}

In this section, we detail the exit head architecture (\S\ref{sec:arch}), the training recipe that makes early layers effective 
(\S\ref{sec:training}), and the depth-wise ``retrieve-rerank'' pipeline (\S\ref{sec:pipeline}).

\subsection{Architecture}
\label{sec:arch}

We start from a standard decoder-only reward model, a pretrained transformer backbone with a scalar reward head attached to the final layer's last-token representation. We augment this with \emph{exit heads}, lightweight modules (MLPs in this work) attached at each intermediate layer $\ell \in \{1, \ldots, L-1\}$. Each exit head reads the last-token hidden state $\mathbf{h}_\ell \in \mathbb{R}^d$ at each layer, applies a two-layer MLP with a residual connection, and produces a scalar score:
\begin{align}
\mathbf{z} &= \text{RMSNorm}(\mathbf{h}_\ell) \nonumber \\
r_\ell(x) &= \mathbf{w}_3^\top \left(\mathbf{W}_2 \, \text{SiLU}\!\left(\mathbf{W}_1 \, \mathbf{z}\right) + \mathbf{z}\right)
\label{eq:exit-head}
\end{align}
where $\mathbf{W}_1 \in \mathbb{R}^{2d \times d}$, $\mathbf{W}_2 \in \mathbb{R}^{d \times 2d}$, and $\mathbf{w}_3 \in \mathbb{R}^d$. Since each head operates on a single vector (not the full sequence), its compute cost is less than 0.1\% of a transformer layer, which gets quickly amortized over the compute savings of filtering via our ``retrieve-rerank'' scheme (\S\ref{sec:pipeline}).

\subsection{Training}
\label{sec:training}

Training comprises two stages, one focusing on the RM backbone and the next
on the training of the scoring heads.
Freezing the backbone lets us train a head at every layer without the heads affecting each other. Concretely:

\paragraph{Stage 1: Backbone.} A pretrained LLM is fine-tuned end-to-end as a reward model, with all parameters unfrozen and a scalar final head trained on preference data. This produces a backbone whose intermediate representations have been shaped by the reward objective at the final layer\footnote{This is similar to the vanilla RM baseline that trains a single output model.}.

\paragraph{Stage 2: Exit heads.} The backbone and original final head are frozen entirely and only the exit heads are trained. We show in \S\ref{sec:training-results} that freezing yields the strongest early-layer exit heads, while also avoiding any increase in peak hardware requirements beyond backbone training.

\paragraph{Loss function.} The standard approach to reward model training is the Bradley-Terry \citep{Bradley1952} (BT) pairwise loss, which trains on chosen/rejected pairs to maximize $\log\sigma(r(y_w) - r(y_l))$. Each pair provides a single binary signal.

We instead train both backbone and exit heads with a \textbf{margin-weighted multi-response BT loss}. The intent is to give every head, and especially the shallow ones, a graded ranking signal over all candidates of a prompt instead of one binary preference; \S\ref{sec:training-results} shows that this matters most at early layers. The loss leverages all $\binom{K}{2}$ pairwise comparisons from $K$ responses simultaneously \citep{Ouyang2022InstructGPT}. Given $K$ responses with gold reward scores $g_1, \ldots, g_K$, we compute the BT loss over all ordered pairs where the gold margin exceeds a threshold $\tau$, weighted by the margin magnitude, analogous to LambdaRank's importance weighting \citep{Burges2006LambdaRank} and the scaled BT loss of \citet{Wang2025HelpSteer2Pref}:
\begin{multline}
\mathcal{L}_{\text{mr-BT}} = \frac{1}{W}\sum_{\substack{i,j:\, g_i - g_j > \tau}} (g_i - g_j)^\alpha \\
\cdot \left(-\log\sigma\bigl(r_\ell(y_i) - r_\ell(y_j)\bigr)\right)
\label{eq:mr-bt-loss}
\end{multline}
where $W = \sum_{i,j:\, g_i - g_j > \tau} (g_i - g_j)^\alpha$ normalizes the weights, $r_\ell$ is the exit head at layer $\ell$, $\alpha{=}1.0$ controls margin upweighting, and $\tau{=}0.5$ filters pairs where the gold RM is uncertain. A ranking of $K$ responses implies up to $\binom{K}{2}$ pairwise preferences. We fit the pairs that pass the margin threshold. A Plackett--Luce objective would instead model the probability of the full ranking \citep{Luce1959,Plackett1975,Zhu2023Kwise}. Because each pair is weighted by its margin, pairs just above $\tau$ contribute little, so $\tau$ only needs to exclude near-ties.

\subsection{Inference: retrieve-rerank Pipeline}
\label{sec:pipeline}

At inference time, we do not run all exit heads. Instead, we adopt a multi-stage ``retrieve-rerank'' pipeline, mirroring the paradigm in information retrieval where a cheap first-stage retriever narrows the candidate set for an expensive reranker \citep{Nogueira2019BERT}, but realized within a single model's depth hierarchy rather than across separate models. The pipeline scores progressively fewer candidates at progressively greater depth:

\begin{enumerate}[leftmargin=1.2em, itemsep=1pt, topsep=2pt]
\item \textbf{Stage 1} (layer $L_{S_1}$): Score all $N$, retain top $K_1$.
\item \textbf{Stage 2} (layer $L_{S_2}$): Score $K_1$ survivors, retain top $K_2$.
\item[$\vdots$]
\item \textbf{Stage $S$} (layer $L_S$): Score $K_{S-1}$ survivors, select winner.
\end{enumerate}

Each stage continues the forward pass from the cached hidden states of the previous stage: layers $[1:L_{S_1}]$ are computed once for all $N$ candidates, then layers $[L_{S_1}{+}1:L_{S_2}]$ are computed only for the $K_1$ survivors, and so on. No layers are recomputed, as early-exiting allows for progressive inference~\citep{laskaridis2020spinn}.

\paragraph{Cost model.} Within a single model, the normalized scoring cost of a pipeline relative to a full forward pass on all $N$ candidates is:
\begin{equation}
C = \sum_{s=1}^{S} \frac{L_s - L_{s-1}}{L} \cdot \frac{K_{s-1}}{N}
\label{eq:cost}
\end{equation}
where $L_0 = 0$, $K_0 = N$, $L$ is the total number of layers. Exit head compute is negligible ($<$0.1\% of one layer) and excluded. This cost model is supported by profiling FLOPs at the lengths of the scored test sequences (median 650--1{,}030 tokens; Appendix~\ref{app:flops}). It counts scoring FLOPs; how savings translate to wall-clock time depends on the serving setup (see Limitations). 

\paragraph{Pipeline optimization.} We jointly optimize the number of stages $S$, layer at each stage $\{L_s\}$, and number of candidates retained $\{K_s\}$ by solving:
\begin{equation}
\max_{\{L_s, K_s\}} \; \text{Reward}(\text{pipeline}) - \lambda \cdot C(\text{pipeline})
\end{equation}
where \textit{Reward} is the mean gold reward of the pipeline's selected candidate on a calibration set ($D_\text{calibration}$). The search enumerates all valid configurations subject to structural constraints (layers ascending, $K$ decreasing, terminal $K{=}1$, bounded stage count) and is tractable because both dimensions are small ($L \leq 36$, $K \leq N$). Sweeping $\lambda$ traces a Pareto frontier from full-model quality to maximum savings. Pseudocode and search-space details are provided in Appendix~\ref{app:search-algorithm}.

\paragraph{Calibration.} We score $N_{\text{cal}}$ responses per calibration prompt through all exit heads once, producing a score matrix. The optimizer operates offline on this
cached
matrix. Configurations transfer to any deployment $N$ by matching retention fractions: e.g., if the optimizer selects top-8 from $N_{\text{cal}}{=}12$, we retain top ${round}(\frac{8}{12} \times N)$ at deployment.

\paragraph{Zero-calibration default.} We observe in \S\ref{sec:inference-results} that optimized pipelines converge to similar structures across settings, yielding a fixed three-stage default as a practical fallback. This heuristic provides a single reasonable operating point without calibration data; the full optimizer is still required to navigate the Pareto frontier and match a specific cost--quality target.

\section{Experimental Setup}
\label{sec:setup}

\subsection{Models and Data}

\paragraph{Models} We train early-exit reward models at five scales across two families: Qwen3 \citep{yang2025qwen3technicalreport} (0.6B, 1.7B, 4B) and Gemma3 \citep{gemmateam2025gemma3technicalreport} (1B, 4B), spanning 26--36 transformer layers. For candidate generation, we use two base policies: Qwen3-1.7B (with thinking mode) and Gemma3-1B (without, as it does not natively support one). In the main setup, each RM is trained, calibrated and evaluated on responses from the same policy, including cross-family settings, e.g., Gemma3 RM trains on Qwen3 responses, then scores Qwen3 BoN candidates; \S\ref{sec:shift} tests generalization to the other policy.

\paragraph{Datasets} We use two preference datasets, namely UltraFeedback (UF)~\citep{Cui2024UltraFeedback} 
and HelpSteer2 (HS2)~\citep{Wang2024HelpSteer2}.
Prompts are split into three disjoint sets (10:1:1 ratio): \textbf{train} for RM training, \textbf{validation} for retrieve-rerank pipeline calibration, \textbf{test} for BoN {evaluation.}

For each prompt, we generate responses and score them with a gold RM (WorldPM-72B \citep{WorldPM2025}; HelpSteer2 and UltraFeedback variants). The gold RM is used in lieu of ground truth, where it is used for labeling purposes simulating human labellers and not otherwise exposed to the RMs. For \texttt{train} and \texttt{validation} we generate $K{=}12$ responses (temperature 1.0, top-$p$ 0.9; the choice of $K$ is ablated in Appendix~\ref{app:robustness}); for \texttt{test} we generate 64 responses to evaluate BoN at \mbox{$N\leq64$}. Thinking tokens are stripped before scoring, as the gold RM was not trained to support rewarding thinking traces. 

\subsection{Evaluation}

We evaluate using Best-of-N selection quality: for each prompt in \texttt{test}, the exit head (or pipeline) selects one candidate from up to 64, and we report the mean gold reward of the selected responses. All results report mean $\pm$ std across three training seeds.
Data splits and candidate generation are fixed. We measure quality gain/loss from our method relative to full-RM scoring, as a fraction of the oracle$-$random range, i.e.\ the gain that selection can deliver on a candidate set. Bradley-Terry rewards have no fixed zero or unit, so we use this range rather than the reward itself as the common scale across configurations:
\begin{equation}
\text{degradation} = \frac{r_{\text{full}} - r_{\text{pipeline}}}{r_{\text{oracle}} - r_{\text{random}}}
\label{eq:degradation}
\end{equation}
where $r_{\text{full}}$ is the mean reward of full-depth RM selection (our baseline), and oracle/random are the max and mean gold reward, as defined below. Here, 1\% of the range is 0.012--0.014 gold reward.

\subsection{Baselines and Comparisons}
\label{sec:baselines}

Our method combines two components: (1) early-exit heads trained with multi-response BT loss, and (2) a ``retrieve-rerank'' depth-wise pipeline. We evaluate against reference points and ablations that each remove one of these two contributions.

\paragraph{Reference points.} \textit{(i)}~\textit{Full RM} (final head scores all 64 candidates, standard BoN), \textit{(ii)}~\textit{Oracle} (highest gold reward), and \textit{(iii)}~\textit{Random} (mean gold reward). This baselines our method against common, best-case and 
degenerate baselines.

\paragraph{Exit heads without pipeline.} A single exit head at a chosen layer scores all $N$ candidates directly, without multi-stage filtering. This removes the pipeline component, testing whether trained exit heads alone suffice.

\paragraph{Pipeline without exit heads.} A retrieve-rerank cascade across multiple models (0.6B $\to$ 1.7B $\to$ 4B) using only their final heads, i.e.\ a model cascade \citep{Chen2023FrugalGPT}. This removes early-exit training, testing whether the pipeline structure alone, relying on model-size diversity, suffices for efficient cascading.

\paragraph{Adaptive exit rules.} Per-input early-exit methods choose one exit depth per prompt instead of filtering candidates across depths. We run two such rules on our trained heads: confidence gating with score-margin, entropy and inter-layer-agreement signals, in the spirit of confidence-based exits \citep{Schuster2022CALM,Fan2024AdaInfer}, and patience-based exiting (PABEE; \citealp{zhou2020bert}), which stops once the top candidate is unchanged for several consecutive layers. This keeps the exit heads and replaces only the pipeline's decision rule.

\begin{figure}[t]
\centering
\includegraphics[width=.95\columnwidth]{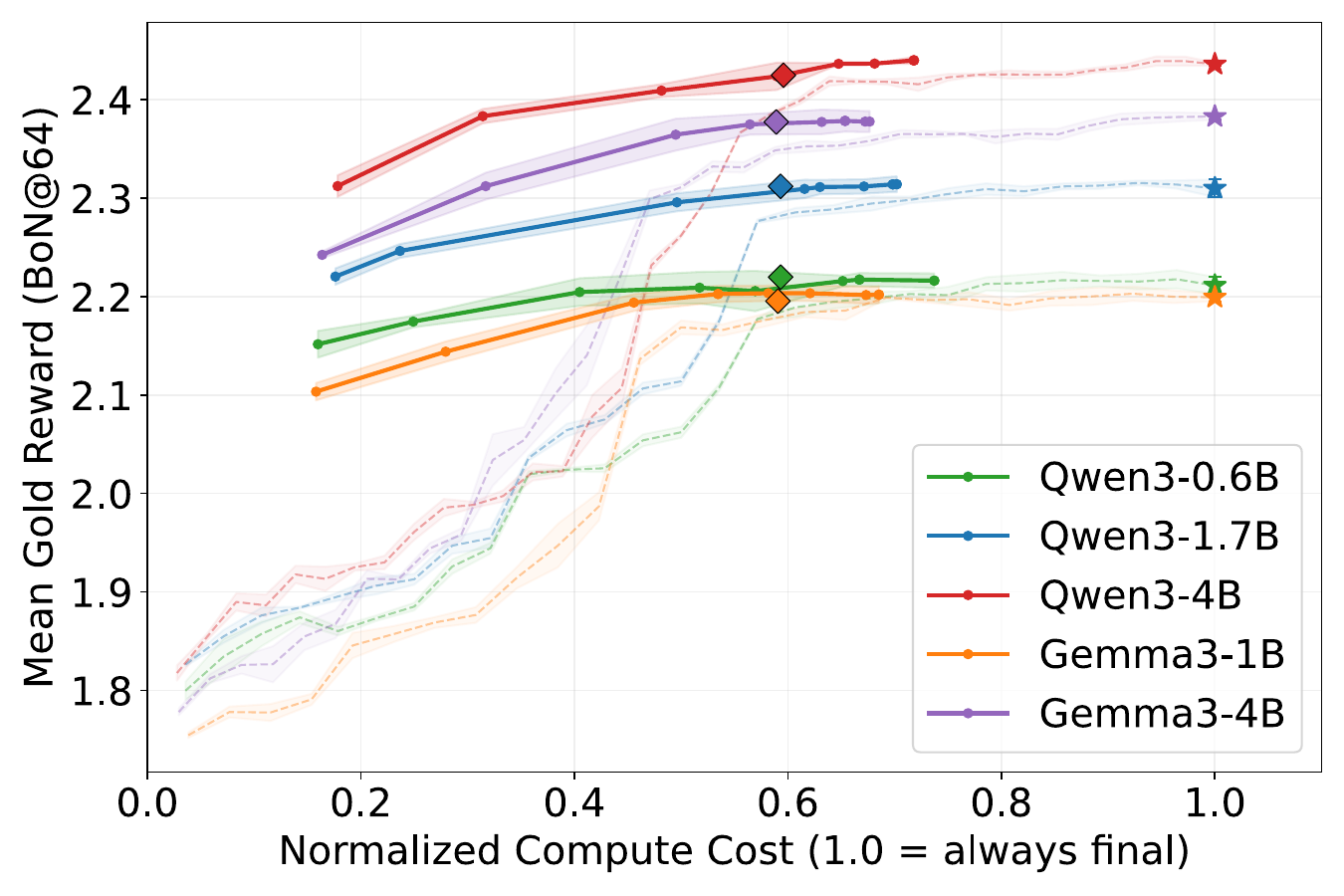}
\vspace{-0.1in}
\caption{Pareto frontiers of performance vs. normalized cost (UltraFeedback, Qwen3-1.7B policy). Shaded bands: $\pm$1 std across 3 seeds. Full 4-cell grid in Appendix~\ref{app:pareto-relative}. Diamonds: heuristic. Stars: full RM.}
\label{fig:pareto-relative}
\vspace{-0.05in}
\end{figure}

\section{Pipeline Performance}
\label{sec:inference-results}

We first evaluate the full depth-wise ``retrieve-rerank'' pipeline (\S\ref{sec:pipeline}). The central question is whether cascading exit heads can cut scoring cost without sacrificing selection quality. Across all 20 configurations (five RMs $\times$ two datasets $\times$ two policies), with the default $\lambda{=}0.05$ the pipeline saves 41\% of RM scoring FLOPs on average, and in 17 of 20 configurations its selection quality is at most 1\% below full depth; changing $\lambda$ moves it along the cost-quality Pareto frontier. The rest of this section unpacks where the savings come from (\S\ref{sec:pareto}), how they extend across model sizes and multi-model cascades (\S\ref{sec:cross-model}), how calibrated pipelines behave under dataset and policy shift (\S\ref{sec:shift}), and how the savings can be reinvested for better selections at a fixed budget (\S\ref{sec:budget}).

\subsection{Cost Savings at Near-Full-Depth Quality}
\label{sec:pareto}

\begin{figure*}[t]
\centering
\includegraphics[width=.89\textwidth]{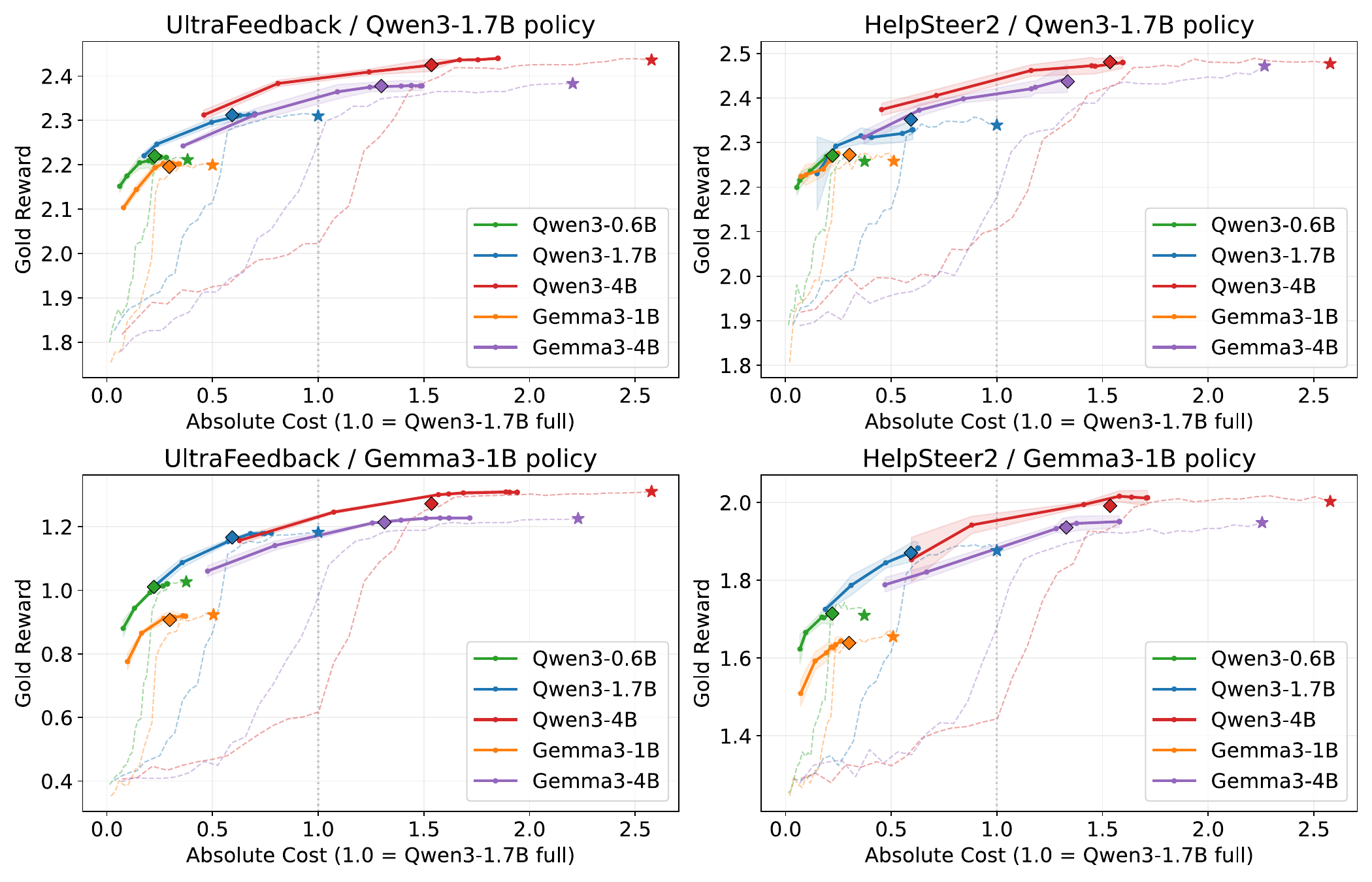}
\vspace{-0.1in}
\caption{Pareto frontiers of performance vs. absolute compute cost (normalized so Qwen3-1.7B full pass = 1.0) across all four cells. Shaded bands: $\pm$1 std across 3 seeds. Solid lines trace optimized pipelines; dashed lines show individual exit layers. Diamonds mark the heuristic; stars mark the full RM. The frontiers form a continuous envelope across model sizes. At any given compute budget, one can read off the best model+pipeline combination.}
\vspace{-0.1in}
\label{fig:pareto}
\end{figure*}

\paragraph{Savings are consistent across models.} Figure~\ref{fig:pareto-relative} plots Pareto frontiers with per-model normalized cost. An interesting regularity emerges: regardless of model family, size, dataset, or policy, each frontier reaches full-RM quality ($<$1\% degradation) at a similar fraction of the model's compute, $55 \pm 7\%$ (mean $\pm$ std over the 19 configurations whose frontier reaches it). This consistency points toward the practicality and transferability of our method's operating points. Table~\ref{tab:operating-points} reports the pipeline at each $\lambda$, averaged across the 20 configurations. At the default $\lambda{=}0.05$, mean degradation is 0.34\% (median 0.22\%), and the pipeline selects the same candidate as full-depth scoring on 75\% of prompts; among the disagreements, the pipeline's pick is the better one in 48\% of cases. $\lambda$ trades savings for quality: $\lambda{=}0.01$ saves 35\% and stays below 1\% degradation in 18 configurations, while $\lambda{=}0.1$ saves 44\% and stays below 1\% in 16.

\begin{table}[t]
\centering
\small
\setlength{\tabcolsep}{2.5pt}
\begin{tabular}{lccccc}
\toprule
$\bm{\lambda}$ & \textbf{Savings} & \textbf{Degr.} & \textbf{$<$1\%} & \textbf{Agreement} & \textbf{Cond.\ win} \\
\midrule
0.001 & 33 $\pm$ 8\% & $-$0.03\% & 19/20 & 82\% & 50\% \\
0.005 & 34 $\pm$ 8\% & $-$0.03\% & 19/20 & 81\% & 49\% \\
0.01 & 35 $\pm$ 8\% & 0.03\% & 18/20 & 80\% & 49\% \\
\textbf{0.05} & \textbf{41 $\pm$ 7\%} & \textbf{0.34\%} & \textbf{17/20} & \textbf{75\%} & \textbf{48\%} \\
0.1 & 44 $\pm$ 8\% & 0.62\% & 16/20 & 71\% & 47\% \\
0.5 & 71 $\pm$ 6\% & 5.16\% & 0/20 & 46\% & 42\% \\
\bottomrule
\end{tabular}
\vspace{-0.1in}
\caption{Pipeline performance with one $\lambda$ shared by all 20 configurations (mean $\pm$ std across configurations, 3 seeds each). Degr.: mean degradation (Eq.~\ref{eq:degradation}; negative if better). $<$1\%: configurations with degradation below 1\%. Agreement: same pick as full depth. Cond.\ win: pipeline's pick is better, among disagreements. Bold: default.}
\label{tab:operating-points}
\end{table}

\paragraph{An alternative zero-calibration heuristic.} The fraction consistency above enables a fixed default that requires no per-deployment calibration: a three-stage pipeline at 38\%/57\%/90\% depth retaining 67\%/25\%/1 candidates. This heuristic saves 41\% of scoring FLOPs at 0.5\% mean degradation, at most 1\% below full depth in 13 of 20 configurations (diamonds in Figures~\ref{fig:pareto-relative}--\ref{fig:pareto}). At the same savings as the default pipeline (41\%), it trails it by only 0.2\% of the oracle$-$random range on average. A practitioner can therefore deploy \tool at this trade-off without any calibration search, at a small cost in optimality, while the optimized pipeline can move along the Pareto frontier through $\lambda$.

\paragraph{Cascading beats single exits at tight budgets.} The depth-wise pipeline's Pareto frontier lies above the single-exit frontier (dashed lines in Figures~\ref{fig:pareto-relative}--\ref{fig:pareto}) wherever it reaches, in all 20 configurations, and the gap widens at aggressive cost targets: at cost 0.35 (Qwen3-1.7B, UF), the best single layer reaches a gold reward of 1.95 versus the pipeline's 2.25. The reason is a division of labor between depths: early layers have high \emph{recall} but poor \emph{precision} (Appendix~\ref{app:recall}), so the pipeline uses shallow layers to filter and delegates the final decision to deeper layers. Qualitatively, early layers discard obviously poor responses on surface cues (formatting, verbosity, structure) while deeper layers adjudicate semantic correctness and the quality of the reasoning written in the response among the shortlist (Appendix~\ref{app:qualitative}).

\paragraph{Adaptive-exit baselines do not close the gap.} At the same quality level, below 1\% mean degradation, each baseline of \S\ref{sec:baselines} gets its best single setting chosen on test data, an upper bound we need because their settings do not transfer from the calibration set (Appendix~\ref{app:exit-rules}). Even so, PABEE saves only 13\% and confidence gating 32\%, against 41\% for the pipeline at the default $\lambda{=}0.05$ (0.34\% mean degradation).

\subsection{Scaling Across Models and Cascades}
\label{sec:cross-model}

\paragraph{Larger models become accessible at smaller-model budgets.} Figure~\ref{fig:pareto} recasts the frontiers on \emph{absolute} compute (normalized so a full Qwen3-1.7B pass = 1.0). The per-model frontiers stitch into a single continuous envelope. Notably, early exiting lets a larger RM undercut a smaller one. Case in point, Qwen3-4B with a pipeline at cost ${\sim}0.81$ outscores Qwen3-1.7B at full depth (gold reward 2.383 vs.\ 2.310).

\begin{figure}[t]
\centering
\begin{subfigure}[t]{.8\columnwidth}
\centering
\adjustbox{clip, trim=0 0 {.5\width} 0}{%
  \includegraphics[width=2\linewidth]{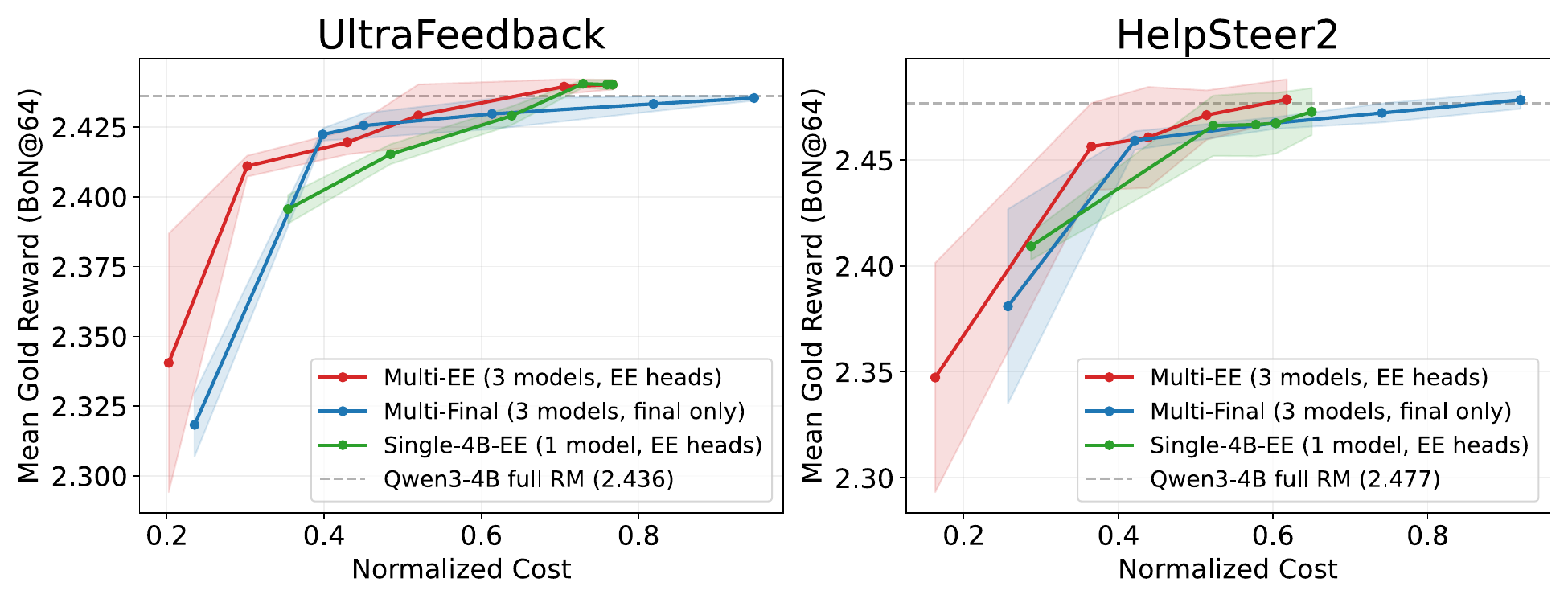}}
\end{subfigure}
\hfill
\begin{subfigure}[t]{.8\columnwidth}
\centering
\adjustbox{clip, trim={.5\width} 0 0 0}{%
  \includegraphics[width=2\linewidth]{figures/multi_model_pareto_grid.pdf}}
\end{subfigure}
\vspace{-0.1in}
\caption{Multi-model cascade Pareto comparison. Multi-EE achieves comparable reward at lower cost than Multi-Final at most operating points. Single-4B-EE converges at quality-focused settings and trails slightly at aggressive cost targets. Shaded bands: $\pm$1 std across 3 seeds.}
\label{fig:multi-model}
\end{figure}

\paragraph{Exit-head cascades are cheaper than multi-model cascades.} A natural alternative to depth cascading is to cascade across separate models. Under an identical pipeline optimizer, we compare three head sets: 
\textit{(i)}~\textit{Multi-EE} (exit heads from 0.6B+1.7B+4B), 
\textit{(ii)}~\textit{Multi-Final} (only those three models' final heads), and 
\textit{(iii)}~\textit{Single-4B-EE} (4B exit heads only). 
Multi-EE reaches Multi-Final's reward at lower cost at most operating points (Figure~\ref{fig:multi-model}): 12--25\% lower cost on UltraFeedback and 3--28\% on HelpSteer2, because fractional-depth filtering is cheaper than routing through separate full forward passes. 
The single-model \emph{Single-4B-EE} converges with both at quality-focused settings while needing only one deployed model, trailing Multi-EE by at most 0.02 gold reward at aggressive targets. This is a modest efficiency price for avoiding multi-model routing.

\subsection{Robustness to Dataset and Policy Shift}
\label{sec:shift}

Each RM is trained on rollouts of one policy and calibrated on prompts of one dataset (\S\ref{sec:setup}). We change one of the two at a time and deploy the RM, its exit heads and its pipelines unchanged (Qwen3-1.7B RM, 3 seeds; per-layer curves under dataset shift in Appendix~\ref{app:ood}): policy shift swaps in the other policy's candidates for the same prompts, dataset shift the other dataset's prompts for the same policy. Pipeline structure, and hence savings, are identical by construction; only quality can change.

\begin{table*}[t]
\centering
\small
\setlength{\tabcolsep}{4pt}
\begin{tabular}{ll c ccc ccc ccc}
\toprule
 & & \textbf{(a) RM}
 & \multicolumn{3}{c}{$\lambda{=}0.005$} & \multicolumn{3}{c}{$\lambda{=}0.05$}
 & \multicolumn{3}{c}{$\lambda{=}0.1$} \\
\cmidrule(lr){4-6}\cmidrule(lr){7-9}\cmidrule(lr){10-12}
\textbf{Data} & \textbf{Policy} & \textbf{shift loss}
 & Sav. & (b) & Total & Sav. & (b) & Total & Sav. & (b) & Total \\
\midrule
UF & Qwen $\to$ Gemma & 6.2\% & 30\% & $-$0.1\% & 6.1\% & 37\% & 1.2\% & 7.4\% & 38\% & 2.1\% & 8.3\% \\
UF & Gemma $\to$ Qwen & 10.7\% & 22\% & 0.2\% & 10.9\% & 26\% & 0.0\% & 10.7\% & 32\% & 0.2\% & 10.9\% \\
HS2 & Qwen $\to$ Gemma & 14.9\% & 40\% & $-$1.1\% & 13.8\% & 45\% & 0.7\% & 15.6\% & 59\% & 0.9\% & 15.8\% \\
HS2 & Gemma $\to$ Qwen & 5.7\% & 37\% & $-$0.8\% & 4.9\% & 37\% & $-$0.8\% & 4.9\% & 39\% & $-$0.5\% & 5.2\% \\
\midrule
HS2 $\to$ UF & Qwen & 24.0\% & 40\% & 0.1\% & 24.1\% & 45\% & $-$0.1\% & 23.9\% & 59\% & 0.4\% & 24.4\% \\
UF $\to$ HS2 & Qwen & 8.7\% & 30\% & 2.1\% & 10.8\% & 37\% & 2.0\% & 10.7\% & 38\% & 2.0\% & 10.7\% \\
\bottomrule
\end{tabular}
\vspace{-0.05in}
\caption{Policy shift (upper rows) and dataset shift (lower rows); $\to$ marks what the RM was trained on and where it is deployed. Losses in \% of the oracle$-$random range, against an RM trained on the deployed policy or dataset at full depth. (a)~The shifted RM at full depth. (b)~The pipeline's degradation on top of (a) (Eq.~\ref{eq:degradation}; negative: better than full depth). Total = (a)+(b). Sav.: scoring-FLOP savings.}
\label{tab:shift}
\vspace{-0.1in}
\end{table*}

Table~\ref{tab:shift} separates the shifted RM's own loss (a) from what the pipeline adds on top (b). Most of the loss belongs to the RM itself: at full depth, the shifted RM falls 5.7--24.0\% of the oracle$-$random range behind (column (a)). This is a property of any RM trained on a single policy or dataset and is orthogonal to our pipeline. We compare what the pipeline adds (column (b)) with its in-domain degradation at the same savings: at the default $\lambda{=}0.05$, (b) averages 0.5\% (at most 2.0\%), against 0.4\% (at most 1.4\%) in domain.

\subsection{Reinvesting Savings into More Candidates}
\label{sec:budget}

So far we have held $N$ fixed and reduced computational cost. However, these savings can instead be spent on additional generations. Because BoN costs $N \times (\text{generation} + \text{scoring})$, reducing scoring frees budget for additional candidates. We define a generation-to-scoring cost ratio $\rho$ (generating a response costs $\rho$ times as much as scoring it with the full RM), and for each budget that would run generation and full-RM scoring we find the best pipeline configuration fitting that budget.

\begin{table}[t]
\centering
\small
\begin{adjustbox}{max width=\columnwidth}    
\begin{tabular}{ccccc}
\toprule
$\bm{N_{\text{budget}}}$ & \textbf{Full RM} & \textbf{Pipe.\ reward} & \textbf{Pipe.\ } $\bm{N}$ & \textbf{Extra} \\
\midrule
16 & 2.153 & 2.167 & 18 & +2 \\
32 & 2.238 & 2.255 & 36 & +4 \\
48 & 2.281 & 2.295 & 54 & +6 \\
56 & 2.297 & 2.310 & 63 & +7 \\
\bottomrule
\end{tabular}
\end{adjustbox}
\vspace{-0.1in}
\caption{Budget allocation ($\rho{=}2.0$, Qwen3-1.7B, UF, 3 seeds; pipeline calibrated at the default $\lambda{=}0.05$). At the budget of full-RM BoN@$N_{\text{budget}}$, the pipeline's scoring savings pay for Pipe.~$N$ candidates (mean over seeds). Rewards: mean gold reward of the selected response.}
\label{tab:budget}
\end{table}

Evaluating on Qwen3-1.7B / UltraFeedback at $\rho{=}2.0$, the pipeline dominates full-RM scoring from $N \geq 12$, and the saved compute buys more candidates as $N$ grows (2 more at $N{=}16$, 7 at $N{=}56$; Table~\ref{tab:budget}), while the reward gain levels off as BoN saturates. At the budget of full-RM BoN at $N{=}56$ (gold reward 2.297), the pipeline generates and scores 63 candidates and reaches 2.310. The effect is naturally sensitive to $\rho$, growing when scoring is relatively expensive and shrinking when generation dominates (Appendix~\ref{app:cost-sensitivity}).

\section{What Makes Early Layers Effective?}
\label{sec:training-results}

The pipeline's savings rest on the premise that early layers are good-enough filters. This is not native, but the product of two training choices, the \textit{loss function} and \textit{a frozen backbone}. We first characterize the per-layer quality profile the pipeline exploits, then isolate each choice's contribution.

\subsection{Per-Layer Reward Curves}
\label{sec:per-layer-reward-curves}

\begin{figure}[t]
\centering
\includegraphics[width=.95\columnwidth]{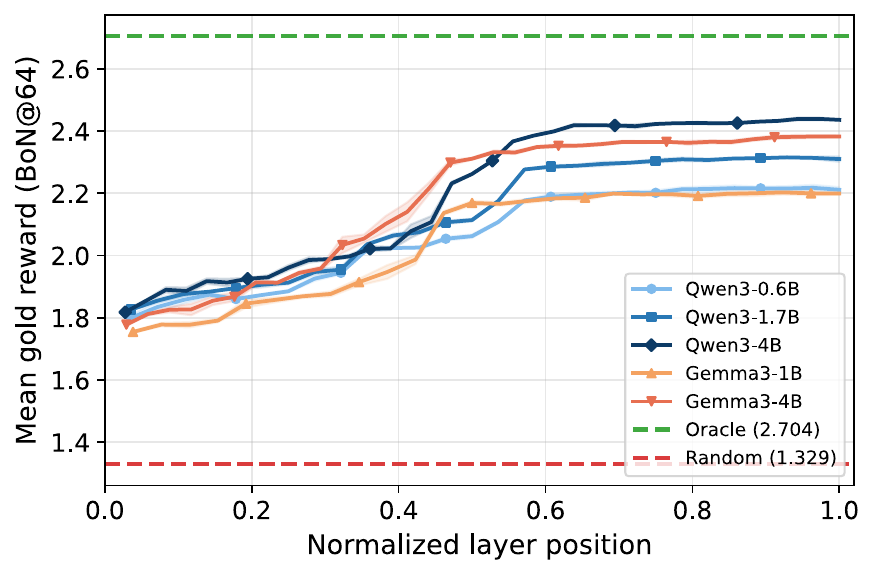}
\vspace{-0.1in}
\caption{Per-layer BoN@64 for five RMs scoring Qwen3-1.7B rollouts on UltraFeedback. All models exhibit a climbing region followed by a plateau (onset at $\sim$60\% depth). Larger models achieve higher absolute quality but the same relative shape.}
\label{fig:per-layer}
\vspace{-0.1in}
\end{figure}

Figure~\ref{fig:per-layer} shows BoN@64 as a function of exit layer for all five RMs on UltraFeedback (Qwen3-1.7B policy). Every model traces the same two-region shape. Even the earliest layers score well above random, and through the \emph{climbing region} (roughly the first 60\% of layers) each added layer improves selection quality. These are the layers that serve as filters in the pipeline. 
Beyond that depth the curve enters a \emph{plateau}, where each layer already captures ${\sim}95\%$ of the final head's gain over random. The shape holds across all four dataset/policy setups (Appendix~\ref{app:per-layer}), with larger RMs reaching higher absolute BoN but the same relative profile. Cross-family pairings show no structural degradation (Qwen3 slightly ahead of Gemma3 at matched size). 
The non-trivial discriminative power of the layers in the climbing region enables them to succeed as high-recall low-precision retrievers;
the next two subsections show which training choices produce it.

\subsection{Loss Function Drives Early-Layer Quality}

\begin{figure}[t]
\centering
\includegraphics[width=.95\columnwidth]{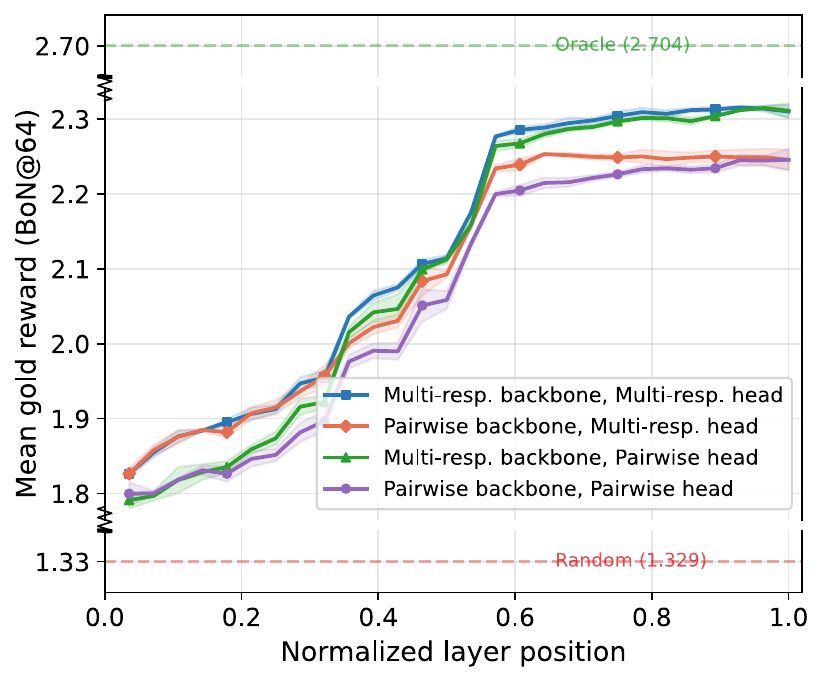}
\vspace{-0.1in}
\caption{2$\times$2 ablation crossing backbone training loss (Multi-response vs.\ Pairwise) with exit-head loss (Multi-response vs.\ Pairwise). UltraFeedback, Qwen3-1.7B policy, Qwen3-1.7B RM. At early layers, multi-response head loss dominates regardless of backbone; at deep layers, backbone loss dominates. Y-axis truncated to better show the exits' range.}
\label{fig:2x2}
\vspace{-0.1in}
\end{figure}

The results above train both backbone and exit heads with the multi-response BT loss. Figure~\ref{fig:2x2} ablates this choice for each component independently (Qwen3-1.7B RM, UltraFeedback). The separation is clean. At early layers ($l \in [1,7]$) the \emph{head} loss accounts for nearly all the difference: multi-response heads beat pairwise heads by ${\sim}4\%$ of the oracle$-$random range, and this holds regardless of how the backbone was trained. At deep layers ($l \in [20,\dots)$) the pattern reverses and the backbone loss dominates. The practical implication is that multi-response training matters precisely where the pipeline filters. One can pair a cheap pairwise backbone with multi-response exit heads, with negligible extra cost on a frozen backbone, and still capture the full early-layer benefit.

\subsection{Freezing the Backbone}
\label{sec:freeze}

\begin{figure}[t]
\centering
\includegraphics[width=.95\columnwidth]{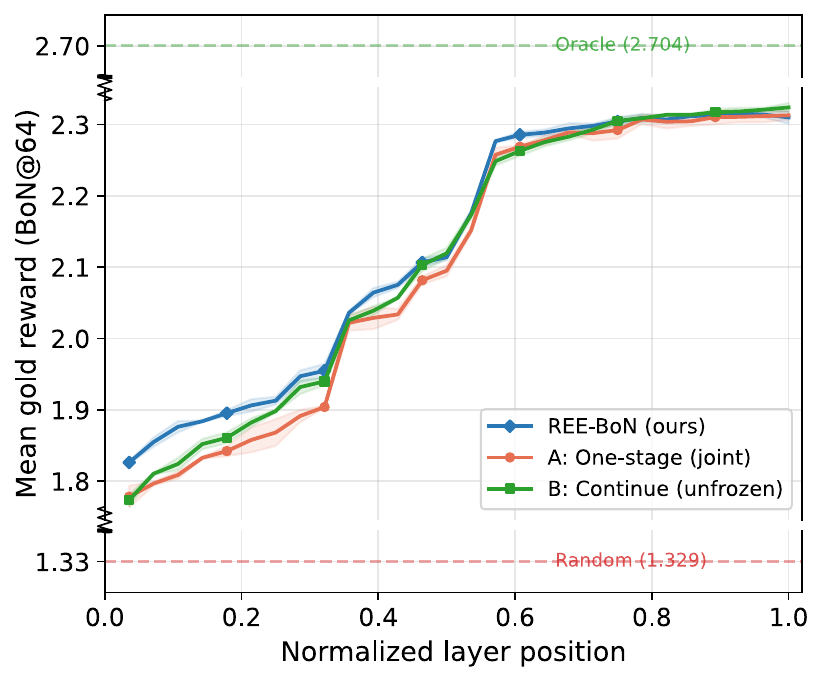}
\vspace{-0.1in}
\caption{Training regime comparison. UltraFeedback, Qwen3-1.7B policy, Qwen3-1.7B RM. Y-axis truncated to better show the exits' range. }
\label{fig:regime}
\vspace{-0.1in}
\end{figure}

Our recipe trains the backbone first, then the exit heads with the backbone frozen (\S\ref{sec:training}). We compare against \textbf{(A)} one-stage joint training of backbone and heads from initialization, and \textbf{(B)} continued training with heads added and the backbone left unfrozen. All three yield functional final-layer heads (gold reward 2.310--2.324 on UF) and differ only at early and middle layers (Figure~\ref{fig:regime}). Our frozen recipe beats variant B at 71--79\% of layers; variant B is ahead mainly at the deepest layers, by small margins (mean +.005 gold reward on UF, +.009 on HS2), and variant A is consistently weakest, lacking the strong Stage-1 backbone initialization. Freezing is also cheaper, with compute peaking in Stage 1, as Stage 2 trains only lightweight MLP heads, therefore making it the clear choice for early exiting (HelpSteer2 consistent, Appendix~\ref{app:regime-hs2}). 
We further confirm robustness to the number of ranked responses per prompt ($K{=}12$ is a robust default) and invariance of the curve shape across evaluation budgets $N \in \{4, 16, 64\}$ (Appendix~\ref{app:robustness}).

\section{Conclusion}
\label{sec:conclusion}

We have shown that a single reward model can serve as an efficient multi-stage scoring pipeline. Shallow exit heads are poor at picking the winner but good at keeping it in a shortlist, and training them on a frozen backbone with a multi-response loss makes them better filters. Filtering with shallow heads and reranking with deep ones keeps full-depth selection quality at a much lower scoring cost, holds up when the dataset or policy changes, and can turn the savings into better selections.

Future directions include rejection sampling at larger candidate scales, composition with adaptive generation-budget methods, and online RLHF.

\section*{Limitations}

\paragraph{Scale.} The largest RM tested is 4B parameters. While the plateau phenomenon is consistent with findings on larger models \citep{Gromov2025Layers}, we have not verified that our pipeline structure and heuristic transfer to larger reward models. The heuristic's depth fractions were chosen from the configurations we report, so its transfer to other model families is also untested.

\paragraph{Evaluation scope.} We evaluate on Best-of-N selection only. Online RLHF (PPO, RLOO) uses reward scores to drive gradient updates, where accuracy requirements are stricter than for filtering, as noisy early-layer scores could introduce bias into policy training. Extending early-exit scoring to online RL remains future work.

\paragraph{Gold RM as proxy.} Our evaluation relies on a gold reward model (WorldPM-72B) as a proxy for human judgment. While 70B-class RMs achieve high agreement with human preferences, human evaluation remains the ground truth. The same WorldPM family also supplies the training labels, so our results show that the pipeline preserves full-depth quality under this shared source of truth; they do not test agreement with an independent judge.

\paragraph{FLOPs, not latency.} We report savings in RM scoring FLOPs. Realized wall-clock savings depend on the serving setup: at batch size 1 pruned candidates simply skip deeper layers, while batched serving must handle shrinking batches, for example with continuous batching or cross-request pipelining. We did not measure end-to-end latency or throughput.

\paragraph{Cost ratio.} The generation-to-scoring cost ratio $\rho$ depends on hardware, batch size, sequence length, and inference framework. We present results parameterized by $\rho$ with sensitivity analysis, but practitioners must measure their own ratio for their deployment.

\paragraph{Scalar reward models only.} Our method applies to RMs that produce a scalar score via a single forward pass. Reasoning reward models that generate chain-of-thought tokens before scoring operate in a fundamentally different computational paradigm where cost is in reasoning-token generation, not model depth.

\section*{Ethical Considerations}

\paragraph{Reward model biases.} Our method inherits any biases present in the underlying reward model and training data. Practitioners should audit for distributional fairness before using early-exit pipelines in production.

\paragraph{Dual use.} Reward models are a core component of LLM alignment. Making RMs cheaper could benefit alignment efforts by enabling more thorough candidate evaluation under fixed budgets. However, the same efficiency gains could lower the cost of training models on objectives that are misaligned with societal values. We believe the net effect is positive.

\paragraph{Environmental impact.} Our work is explicitly motivated by reducing the computational cost of reward scoring, which may reduce energy use at inference time. Training the exit heads adds modest overhead (a single epoch of reward head fine-tuning), which is amortized over the lifetime of the deployed model.

\section*{Acknowledgments}

We thank the anonymous reviewers and the area chair for their constructive feedback, which helped improve this paper.

\bibliography{references}

\clearpage
\appendix

\part{Supplementary Material}
\parttoc

In this Appendix, we provide additional details for our methodology along with additional pointers for enhancing reproducibility (\S\ref{app:methodological-details}). Additionally, we go through supplementary results that showcase the behavior of our method across different setups (\S\ref{app:additional-results}) that were not covered in the main body of the manuscript and provide qualitative examples of the trained reward models' selections (\S\ref{app:qualitative}). Last, we compare the pipeline with alternatives that use the same trained exit heads, namely per-prompt adaptive exit rules and multi-exit ensembling (\S\ref{app:negative-results}).

\section{Methodological Details}
\label{app:methodological-details}

In this section, we complement the methodology of \tool with additional details, including preference pair generation, algorithmic representation of our method and hyperparameters used across stages.

\subsection{Pipeline Optimization Algorithm}
\label{app:search-algorithm}

Here, we provide a detailed algorithmic outline of the pipeline optimisation of \tool, initially introduced in \S\ref{sec:pipeline}, and shown in Algorithm~\ref{alg:pipeline-search}.

Starting from the trivial baseline of scoring all N candidates through the full model, the optimizer performs a depth-first search over multi-stage configurations. Each stage is defined by a layer and a retention count, constrained so that layers strictly increase and retention counts strictly decrease across stages, terminating at a single winner. For each candidate extension, the search simulates \emph{filtering} (ranking survivors by that layer's exit-head scores and keeping the top-K) then \emph{evaluates an objective} balancing selection quality (mean gold reward of the pipeline's final picks) against normalized compute cost (the sum, over stages, of the fraction of layers spanned times the fraction of candidates processed). A tradeoff parameter $\lambda$ controls this balance; sweeping it traces a Pareto frontier from maximum quality to maximum savings. 
Results for different values of $\lambda$ are summarized in Table~\ref{tab:operating-points}; \S\ref{app:pipelines} shows the selected pipelines.

\begin{algorithm}[h]
\caption{Pipeline Optimization}
\label{alg:pipeline-search}
\small
\setlength{\algorithmicindent}{1.2em}
\begin{spacing}{0.92}
\begin{algorithmic}[1]
\REQUIRE Score matrix $\mathbf{S}^{P \times N \times L}$, gold rewards $\mathbf{G}^{P \times N}$, tradeoff $\lambda$, max stages $S_{\max}$
\ENSURE Pipeline $\mathcal{P}^* = [(l_1, K_1), \ldots, (l_m, 1)]$
\STATE Initialize: $\mathcal{P}^* \leftarrow [(\text{final}, 1)]$, $\text{obj}^* \leftarrow \text{Eval}(\mathcal{P}^*) - \lambda \cdot C(\mathcal{P}^*)$
\STATE \textsc{Search}($[\,]$, all candidates, $l_{\text{prev}}{=}0$, $K_{\text{prev}}{=}N$, $0$)
\STATE \rule{0pt}{1.2em}\textbf{procedure} \textsc{Search}($\mathcal{P}$, $\mathcal{C}$, $l_{\text{prev}}$, $K_{\text{prev}}$, depth)
\IF{depth $= S_{\max}$}
\RETURN
\ENDIF
\FOR{layer $l > l_{\text{prev}}$, \; $K < K_{\text{prev}}$}
\STATE $\mathcal{C}' \leftarrow \text{TopK}(\mathbf{S}[\cdot, \mathcal{C}, l],\; K)$ \hfill $\triangleright$ filter
\STATE obj $\leftarrow \text{Eval}(\mathcal{P} \cup [(l,K)]) - \lambda \cdot C(\mathcal{P} \cup [(l,K)])$
\IF{obj $> \text{obj}^*$}
\STATE update $\mathcal{P}^*$, $\text{obj}^*$
\ENDIF
\IF{$K > 1$}
\STATE \textsc{Search}($\mathcal{P} \cup [(l,K)]$, $\mathcal{C}'$, $l$, $K$, depth$+1$)
\ENDIF
\ENDFOR
\end{algorithmic}
\end{spacing}
\vspace{-0.3em}
{\footnotesize Eval$(\mathcal{P})$: apply pipeline stages sequentially, return mean gold reward of selected candidates (the top-1 of the last stage's head). At deployment the retention counts are rescaled to $N$; the last stage still returns its top-1, a stage whose rescaled count does not shrink the shortlist is skipped, and ties go to the lowest candidate index. $C(\mathcal{P})$: Eq.~\ref{eq:cost}. Constraints: layers ascending, $K$ decreasing, terminal $K{=}1$.}
\end{algorithm}

\subsection{Training Setup \& Hyperparameters}
\label{app:hyperparams}

Next, we document in greater detail the training and hyperparameter setup we used for producing the results of our evaluation.

\paragraph{Frameworks \& Hardware.} All models were trained with TRL with DeepSpeed ZeRO-2 on 8$\times$H100 GPUs in \texttt{bfloat16} with gradient checkpointing enabled. We used three seeds, namely $42, 43, 44$. Qwen3 RMs were trained with \texttt{transformers} 4.53; Gemma3 RMs require \texttt{transformers}~$\geq$5 for their sequence-classification head and were trained with 5.7.

\paragraph{Optimizer.} AdamW ($\beta_1{=}0.9$, $\beta_2{=}0.999$, $\epsilon{=}10^{-8}$) for both stages.

\paragraph{Checkpoints.} Both stages train for exactly 1 epoch with no early stopping; we always use the final checkpoint.

\paragraph{Response generation.} Responses are generated with vLLM using temperature 1.0, top-$p$ 0.9, top-$k$ 20, max tokens 2048. Training responses ($K{=}12$) use seed 123; BoN test responses ($N{=}64$) merge 16 generations with seed 123 (12 + 4) and 48 with seed 789. The additional generations for the $K{=}16$ ablation use seed 456.

\paragraph{Pipeline search space.} Maximum 4 stages; layers must be ascending; $K$ must be decreasing with terminal $K{=}1$. Candidate stages are the exit heads at layers $1, \ldots, L{-}1$ and the final head. Search is exhaustive within these constraints (Algorithm~\ref{alg:pipeline-search}).

\paragraph{Multi-EE cascade.} For the multi-model comparison (Figure~\ref{fig:multi-model}), we use exit heads from Qwen3-0.6B, 1.7B, and 4B jointly. To keep search tractable, we subsample every 4th layer for 28-layer models (layers 3, 7, 11, 15, 19, 23, 27 + final) and every 5th for 36-layer models (layers 4, 9, 14, 19, 24, 29, 34 + final), yielding 24 candidate heads total. The search allows at most 3 stages, with $K \in \{1, 2, 3, 4, 6, 8, 9, 10\}$ at $N_{\text{cal}}{=}12$.

\paragraph{Stage 1: Backbone training.}
We trained standard reward models with the margin-weighted multi-response BT loss (Eq.~\ref{eq:mr-bt-loss}). Table~\ref{tab:hyper-stage1} lists selected hyperparameters for training of the RM backbone (nominal model + last head).

\begin{table}[h]
\centering
\small
\begin{adjustbox}{max width=.8\columnwidth}
\begin{tabular}{ll}
\toprule
\textbf{Parameter} & \textbf{Value} \\
\midrule
Learning rate & $5 \times 10^{-6}$ \\
LR scheduler & constant \\
Epochs & 1 \\
Per-device batch size & 1 \\
Gradient accumulation & 4 \\
Effective batch size & 32 \\
Weight decay & 0.01 \\
Center rewards coeff. & 0.01 \\
Max sequence length & 4096 \\
Margin weight ($\alpha$) & 1.0 \\
Min margin & 0.5 \\
Margin floor ($\epsilon$) & 0.01 \\
Responses per prompt ($K$) & 12 \\
\bottomrule
\end{tabular}
\end{adjustbox}
\vspace{-0.1in}
\caption{Stage 1 (backbone) hyperparameters.}
\label{tab:hyper-stage1}
\vspace{-0.1in}
\end{table}

\paragraph{Stage 2: Exit head training.}
The backbone is frozen entirely, including the final reward head. MLP exit heads are placed after transformer blocks $1, \ldots, L{-}1$ (the last block feeds the final reward head). The final projection ($d{\to}1$) and input RMSNorm of each head are initialized from the backbone's final reward head and final layer norm respectively; the intermediate MLP layers (W$_1$, W$_2$) are initialized randomly. Training uses the same multi-response BT loss applied independently per head.

\begin{table}[h]
\centering
\small
\begin{adjustbox}{max width=.8\columnwidth}
\begin{tabular}{ll}
\toprule
\textbf{Parameter} & \textbf{Value} \\
\midrule
Learning rate & $1 \times 10^{-4}$ \\
LR scheduler & cosine \\
Epochs & 1 \\
Per-device batch size & 1 \\
Gradient accumulation & 4 \\
Effective batch size & 32 \\
Weight decay & 0.01 \\
Center rewards coeff. & 0.01 \\
Freeze backbone & yes \\
Head type & MLP \\
Head placement & all layers \\
Init from final head & yes \\
Main loss weight & 0.0 \\
Multi-head loss weight & 1.0 \\
Margin weight ($\alpha$) & 1.0 \\
Min margin & 0.5 \\
Margin floor ($\epsilon$) & 0.01 \\
Responses per prompt ($K$) & 12 \\
\bottomrule
\end{tabular}
\end{adjustbox}
\vspace{-0.1in}
\caption{Stage 2 (exit head) hyperparameters.}
\label{tab:hyper-stage2}
\vspace{-0.1in}
\end{table}

\paragraph{MLP head architecture \& initialization.} Each exit head processes the last-token hidden state: RMSNorm $\to$ Linear($d \to 2d$) $\to$ SiLU $\to$ Linear($2d \to d$) $\to$ residual add $\to$ Linear($d \to 1$), where $d$ is the model's hidden dimension. The intermediate MLP layers are initialized randomly; the final $d{\to}1$ projection and input RMSNorm are copied from the backbone's trained final head and layer norm. Total parameter overhead for all 27 exit heads on Qwen3-1.7B ($L{=}28$; $d{=}2048$, $4d^2{+}2d$ = 16.8M parameters per head): 453M parameters (26.3\% of the backbone). These heads are all trained so as to provide the full search space, but can be heuristically pruned for reducing the training overhead of Stage 2. At inference time, a deployed pipeline uses only the heads of its stages: at most four (67M parameters, 3.9\% of the backbone), and three for the zero-calibration heuristic (50M, 2.9\%). Each of these has an overhead of $\leq0.5$\% of the total RM inference cost, with the upside of exiting early. 

\paragraph{Training time.} Measured on 8$\times$H100 with DeepSpeed ZeRO-2. Table~\ref{tab:training-time} reports wall-clock times for HelpSteer2 (4,621 prompts); UltraFeedback (27,686 prompts) scales approximately 6$\times$ longer.

\begin{table}[t]
\centering
\small
\begin{adjustbox}{max width=\columnwidth}
\begin{tabular}{lccc}
\toprule
\textbf{Model} & \textbf{Stage 1} & \textbf{Stage 2} & \textbf{Total} \\
\midrule
Qwen3-0.6B (28L) & 11 min & 3 min & 14 min \\
Qwen3-1.7B (28L) & 15 min & 4 min & 19 min \\
Qwen3-4B (36L) & 34 min & 10 min & 44 min \\
\bottomrule
\end{tabular}
\end{adjustbox}
\vspace{-0.1in}
\caption{Training wall-clock time on HelpSteer2 (8$\times$H100). Stage 2 is faster because only MLP head parameters are optimized; the frozen backbone forward pass dominates both stages equally, but Stage 2 has no backbone gradient computation.}
\label{tab:training-time}
\vspace{-0.1in}
\end{table}

\paragraph{Data.} Both stages train on the same \texttt{train} split. For each prompt, all $K{=}12$ responses are presented to the loss simultaneously. Thinking tokens (\texttt{<think>...</think>}) are stripped prior to tokenization. Sequences exceeding 4096 tokens after tokenization are dropped entirely (affects $<$2\% of examples).

\subsection{Compute Cost Validation}
\label{app:flops}

We validate the cost model (Eq.~\ref{eq:cost}) via FLOP profiling using \texttt{torch.utils.flop\_counter.FlopCounterMode}, which counts actual dispatched tensor operations at runtime. All models are profiled on GPU with \texttt{attn\_implementation="eager"} (ensuring attention matmuls route through counted ops), batch size 1, inference only.

\paragraph{Per-layer uniformity.} Table~\ref{tab:flops} reports per-layer and per-head FLOPs at a sequence length of 1536 tokens; uniformity holds at every length we profiled (642--1536 tokens). Across all five RMs, per-layer FLOPs are \emph{exactly} uniform (0\% standard deviation across layers), confirming that normalized layer fraction equals normalized FLOPs. Exit head FLOPs are constant across layers and negligible ($<$0.07\% of one backbone layer).

\begin{table}[h]
\centering
\small
\begin{adjustbox}{max width=\columnwidth}
\begin{tabular}{lcp{1.2cm}p{1.2cm}c}
\toprule
\textbf{Model} & \textbf{Layers} & \textbf{Layer (GFLOPs)} & \textbf{Head (MFLOPs)} & \textbf{Head/Layer} \\
\midrule
Qwen3-0.6B & 28 & 67.6 & 8.4 & 0.012\% \\
Qwen3-1.7B & 28 & 173.9 & 33.6 & 0.019\% \\
Qwen3-4B & 36 & 348.7 & 52.4 & 0.015\% \\
Gemma3-1B & 26 & 92.1 & 10.6 & 0.012\% \\
Gemma3-4B & 34 & 309.2 & 52.4 & 0.017\% \\
\bottomrule
\end{tabular}
\end{adjustbox}
\caption{FLOP profiling at seq\_len = 1536. Per-layer FLOPs are perfectly uniform within each model; head FLOPs are negligible.}
\label{tab:flops}
\end{table}

\paragraph{Cross-model FLOP ratios.} For absolute cost comparisons across models (Figure~\ref{fig:pareto}), we normalize by total forward-pass FLOPs rather than parameter count. Per cell, we use each RM's expected forward-pass FLOPs over the scored test sequences of that cell (its own tokenizer and chat template; median 648--1{,}029 tokens across cells and tokenizers); relative to Qwen3-1.7B these are Qwen3-0.6B 0.37--0.38$\times$, Qwen3-4B 2.58$\times$, Gemma3-1B 0.50--0.51$\times$ and Gemma3-4B 2.20--2.27$\times$ across the four cells.

\paragraph{Test-set validation.} Because every layer costs the same at any length, normalized cost at exit layer $L$ equals $L/N_{\text{layers}}$ for every scored sequence, whatever its length. The quadratic attention component is identical at every layer, so it does not affect the cost \emph{ratio} between exit points.

\section{Additional Results}
\label{app:additional-results}

In the following section, we go through supplemental results and experiments that complement the main results of the paper and shed additional light on the behavior of the models that resulted from \tool training.

\subsection{Recall and Precision of Exit Heads}
\label{app:recall}

Taking the zero-calibration heuristic pipeline as an example, Table~\ref{tab:recall} measures how often an exit head at each of its three stage layers keeps the full-depth RM's selected candidate among its top-$K$ (recall) and how often it ranks that candidate first (top-1). The top-43 and top-16 columns are the heuristic's retention at its first two stages. At 38\% depth the head ranks the full-depth choice first for only 11--21\% of prompts, but keeps it among its top 43 of 64 for 93--97\%; by 57\% depth its top 16 keep it for 94--98\%. This is the regime the retrieve-rerank pipeline exploits: shallow heads filter, deep heads decide.

\begin{table}[h]
\centering
\small
\begin{adjustbox}{max width=\columnwidth}
\begin{tabular}{llccc}
\toprule
\textbf{Dataset / policy} & \textbf{Depth} & \textbf{Top-43} & \textbf{Top-16} & \textbf{Top-1} \\
\midrule
\multirow{3}{*}{UF / Qwen3-1.7B} & 38\% (L11) & 96.3\% & 77.6\% & 20.4\% \\
 & 57\% (L16) & 98.7\% & 94.5\% & 44.7\% \\
 & 90\% (L25) & 99.6\% & 98.9\% & 77.0\% \\
\midrule
\multirow{3}{*}{HS2 / Qwen3-1.7B} & 38\% (L11) & 97.3\% & 75.6\% & 14.8\% \\
 & 57\% (L16) & 99.7\% & 97.4\% & 46.8\% \\
 & 90\% (L25) & 100.0\% & 99.8\% & 70.9\% \\
\midrule
\multirow{3}{*}{UF / Gemma3-1B} & 38\% (L11) & 94.5\% & 70.7\% & 21.4\% \\
 & 57\% (L16) & 99.5\% & 97.1\% & 53.5\% \\
 & 90\% (L25) & 99.9\% & 99.5\% & 83.6\% \\
\midrule
\multirow{3}{*}{HS2 / Gemma3-1B} & 38\% (L11) & 93.3\% & 66.7\% & 11.1\% \\
 & 57\% (L16) & 99.9\% & 98.0\% & 46.3\% \\
 & 90\% (L25) & 99.9\% & 99.9\% & 77.4\% \\
\midrule
\multicolumn{2}{l}{Chance} & 67.2\% & 25.0\% & 1.6\% \\
\bottomrule
\end{tabular}
\end{adjustbox}
\caption{Share of prompts whose full-depth selection is among an exit head's top-$K$ of the 64 BoN candidates (Qwen3-1.7B RM, mean over 3 seeds), at the stage layers of the heuristic pipeline, taken as an example.}
\label{tab:recall}
\end{table}

\subsection{Full Per-Layer Results}
\label{app:per-layer}

\begin{figure*}[h]
\centering
\begin{subfigure}{0.32\textwidth}
  \centering
  \includegraphics[width=\textwidth]{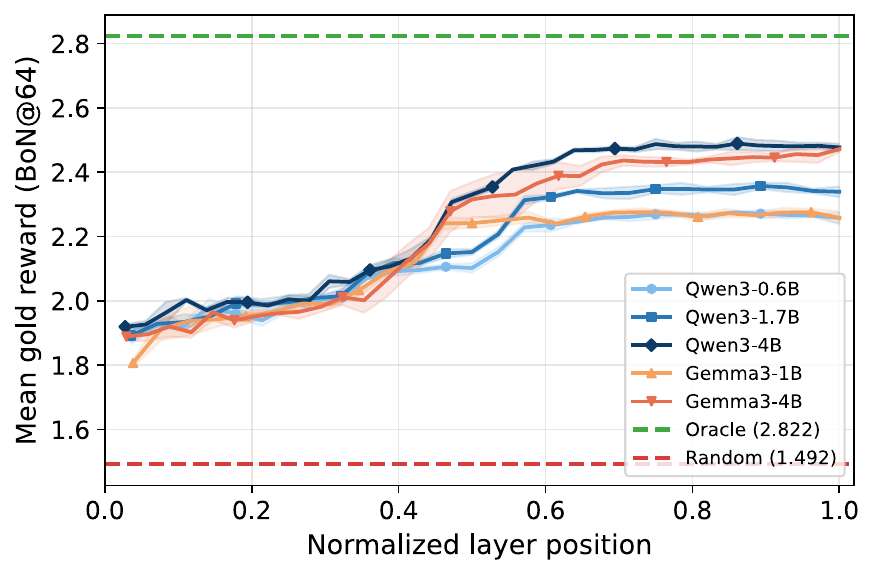}
  \caption{HS2 / Qwen3-1.7B}
  \label{fig:per-layer-hs2-qwen1.7b}
\end{subfigure}\hfill
\begin{subfigure}{0.32\textwidth}
  \centering
  \includegraphics[width=\textwidth]{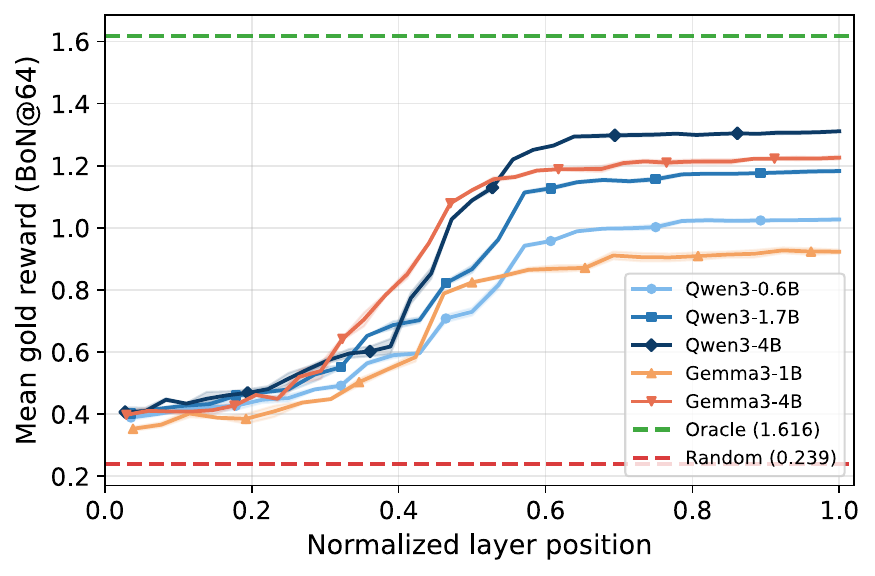}
  \caption{UF / Gemma3-1B}
  \label{fig:per-layer-uf-gemma1b}
\end{subfigure}\hfill
\begin{subfigure}{0.32\textwidth}
  \centering
  \includegraphics[width=\textwidth]{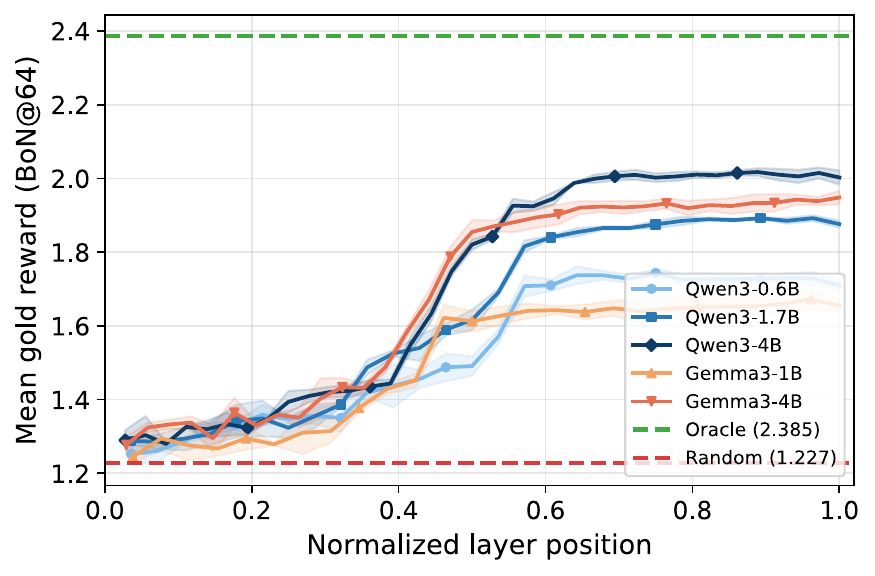}
  \caption{HS2 / Gemma3-1B}
  \label{fig:per-layer-hs2-gemma1b}
\end{subfigure}
\vspace{-0.1in}
\caption{Per-layer BoN@64 for the remaining three cells. The ``climbing-then-plateau'' pattern is consistent across all settings.}
\label{fig:app-per-layer}
\end{figure*}

We provide the complete picture of the per-layer BoN@64 curves, complementing Figure~\ref{fig:per-layer} of the main text. Figure~\ref{fig:app-per-layer} follows the pattern of \S\ref{sec:per-layer-reward-curves}, with the \emph{climbing} and \emph{plateau} regions clearly visible across setups.

\subsection{Training Regime (HelpSteer2)}
\label{app:regime-hs2} 

\begin{figure}[h]
\centering
\includegraphics[width=.95\columnwidth]{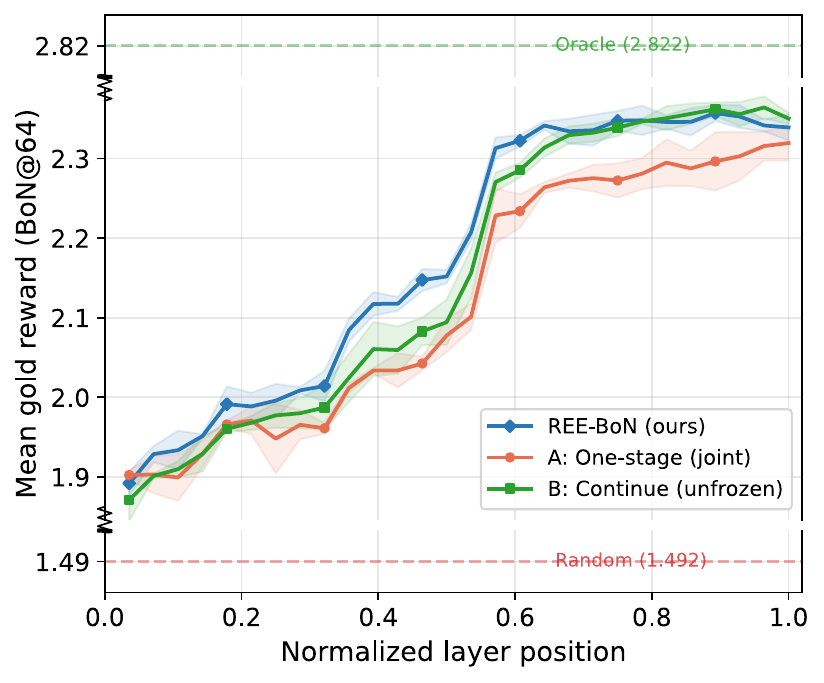}
\vspace{-0.1in}
\caption{Training regime comparison: HelpSteer2, Qwen3-1.7B. Same pattern as UltraFeedback (Figure~\ref{fig:regime}).}
\label{fig:app-regime-hs2}
\vspace{-0.1in}
\end{figure}

Next, we complement the results of Figure~\ref{fig:regime} with the evaluation results when Qwen3-1.7B is trained on HelpSteer2. 
Figure~\ref{fig:app-regime-hs2} confirms the regime comparison on HelpSteer2 and shows the same pattern beyond a single dataset.

\subsection{Relative-Cost Pareto Frontiers}
\label{app:pareto-relative}

Subsequently, we showcase the rest of the results from the Pareto frontier comparison depicted in Figure~\ref{fig:pareto-relative} for Qwen3-1.7B on HelpSteer2, and Gemma3-1B on UltraFeedback and HelpSteer2.

\begin{figure*}[h]
\centering
\begin{subfigure}{0.32\textwidth}
  \centering
  \includegraphics[width=\textwidth]{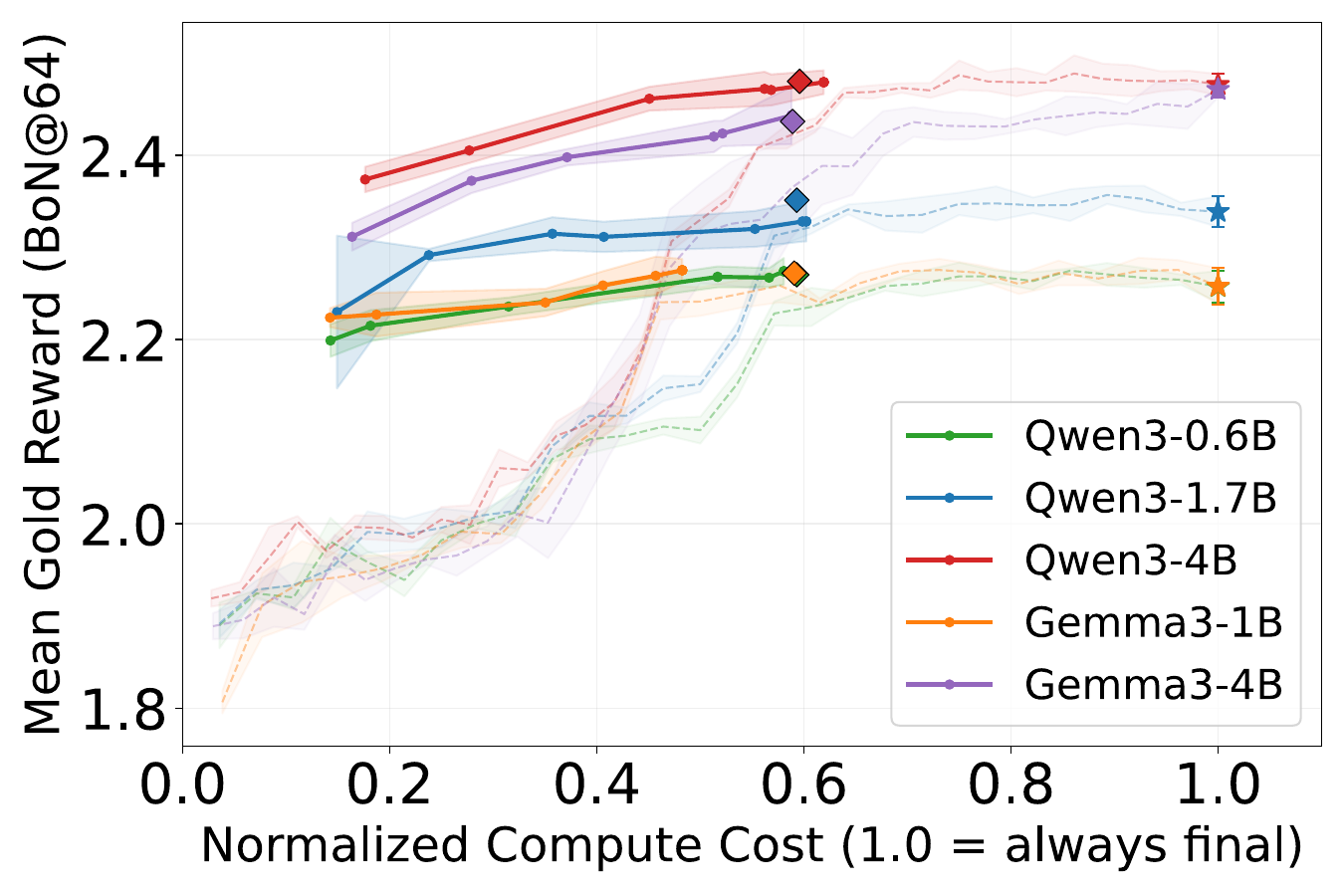}
  \caption{HS2 / Qwen3-1.7B}
  \label{fig:pareto-hs2-qwen1.7b}
\end{subfigure}\hfill
\begin{subfigure}{0.32\textwidth}
  \centering
  \includegraphics[width=\textwidth]{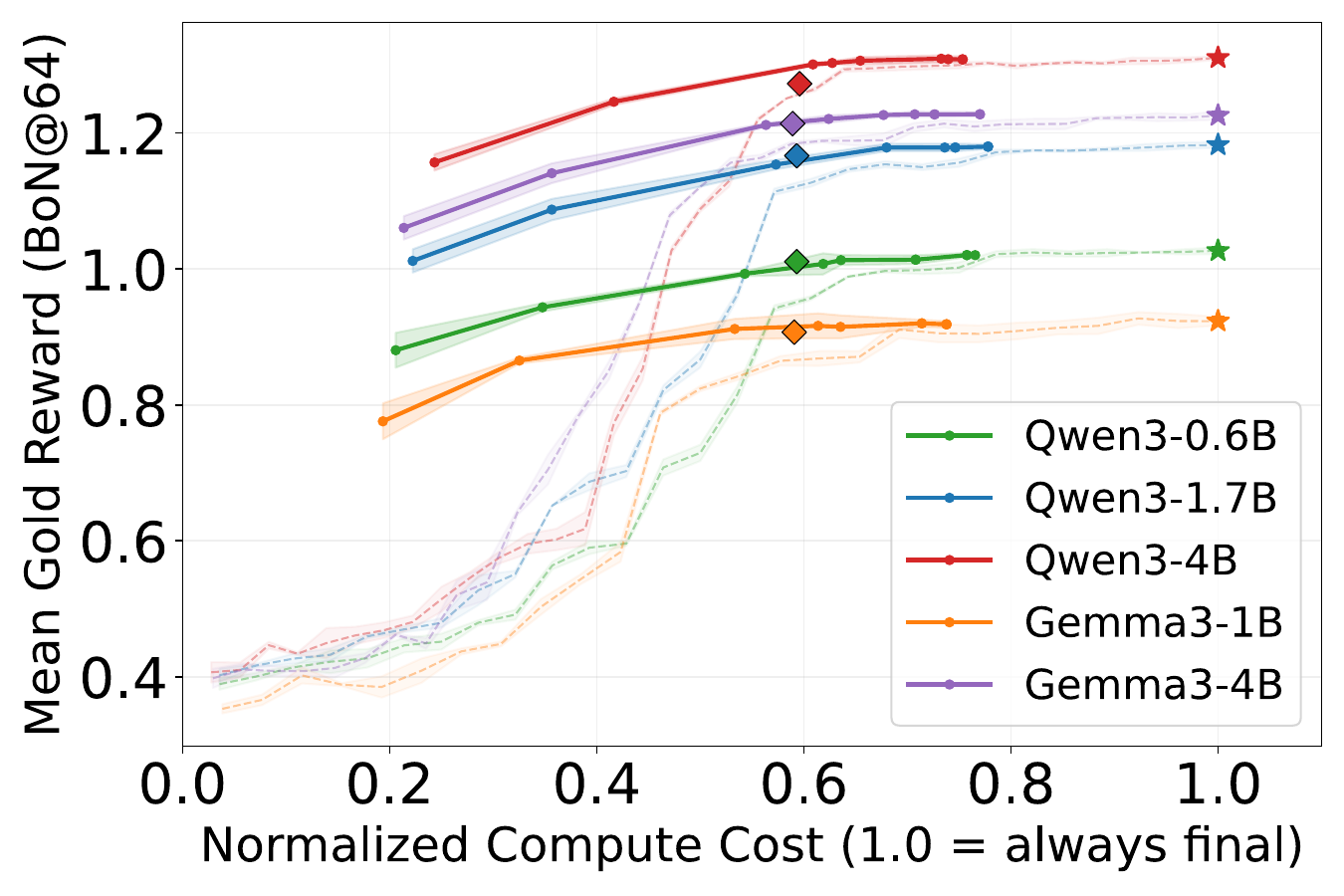}
  \caption{UF / Gemma3-1B}
  \label{fig:pareto-uf-gemma1b}
\end{subfigure}\hfill
\begin{subfigure}{0.32\textwidth}
  \centering
  \includegraphics[width=\textwidth]{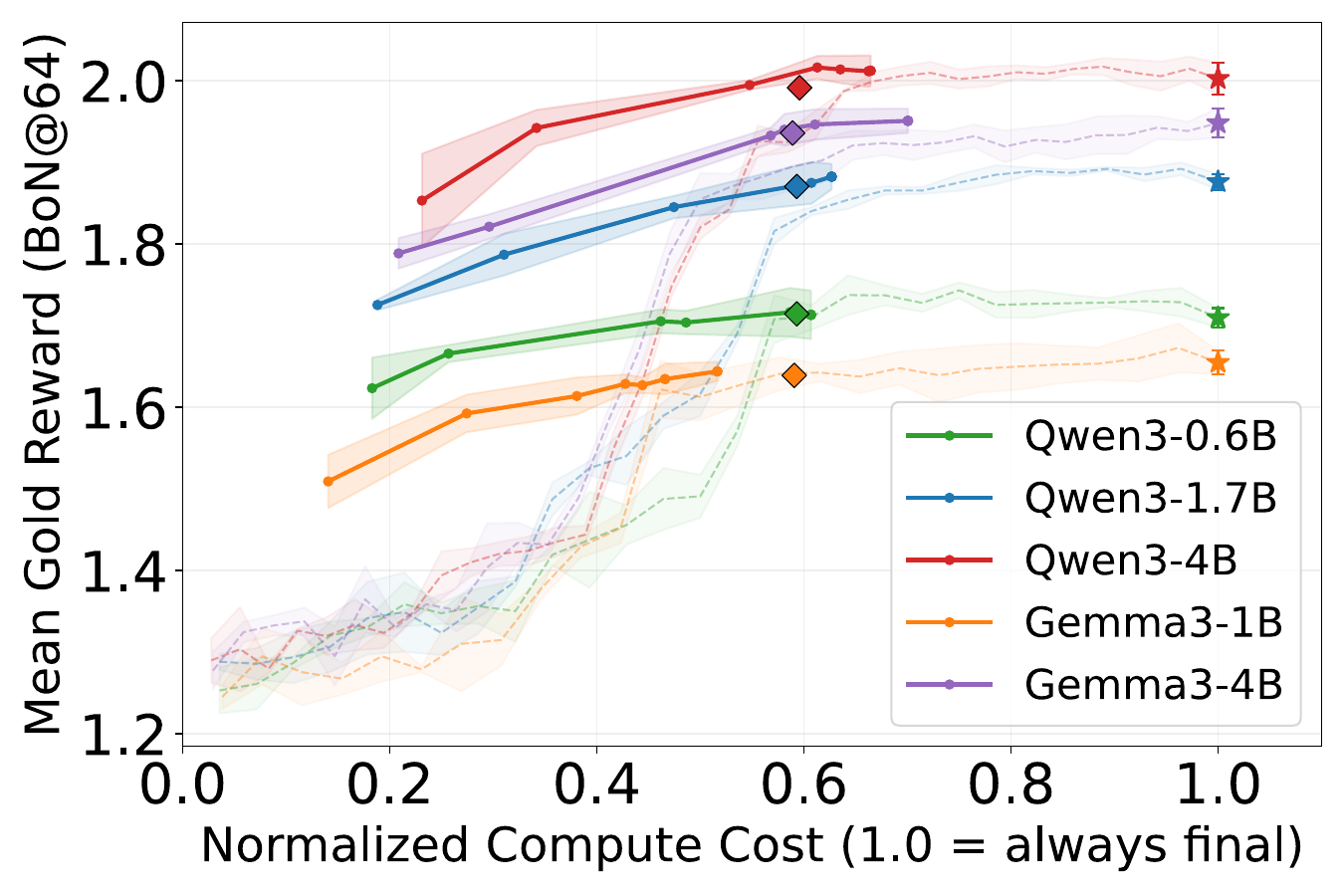}
  \caption{HS2 / Gemma3-1B}
  \label{fig:pareto-hs2-gemma1b}
\end{subfigure}
\caption{Per-model normalized Pareto frontiers for the remaining three cells. Shaded bands: $\pm$1 std across 3 seeds. Diamonds: heuristic. The same fraction consistency (${\sim}55\%$ of compute to come within 1\%) holds across the configurations that reach it.}
\label{fig:app-pareto}
\end{figure*}

\subsection{Robustness and Sensitivity}
\label{app:robustness}

In this section, we analyze the model's sensitivity to various hyperparameters found in the evaluation process, namely: \textit{(i)}~the number of generated responses by the policy model ($N$), \textit{(ii)}~the number of ranked responses ($K$), \textit{(iii)}~the size of calibration set ($D_\text{calibration}$), \textit{(iv)}~the cost ratio parameter ($\rho$) and \textit{(v)}~the pairwise sensitivity threshold ($\tau_\text{pair}$).

\subsubsection{Number of Generated Responses ($N$)}
\label{app:n-invariance}

\begin{figure*}[h]
\centering
\begin{subfigure}{0.32\textwidth}
  \centering
  \includegraphics[width=\textwidth]{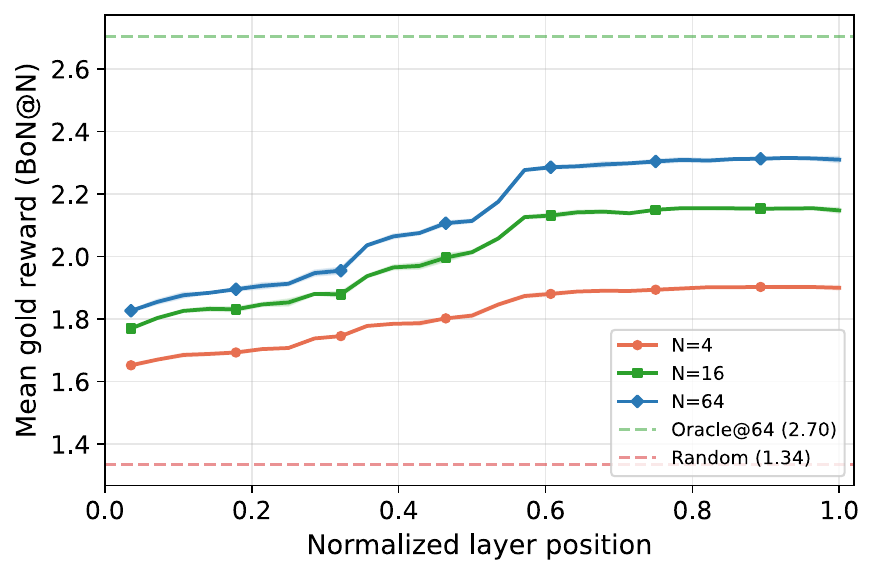}
  \caption{Qwen3-1.7B}
  \label{fig:n-uf-1.7b}
\end{subfigure}\hfill
\begin{subfigure}{0.32\textwidth}
  \centering
  \includegraphics[width=\textwidth]{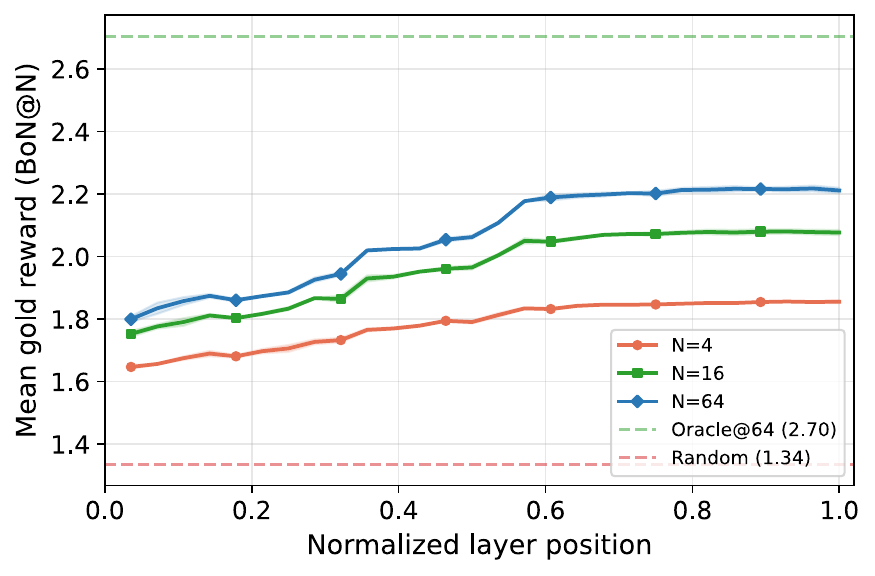}
  \caption{Qwen3-0.6B}
  \label{fig:n-uf-06b}
\end{subfigure}\hfill
\begin{subfigure}{0.32\textwidth}
  \centering
  \includegraphics[width=\textwidth]{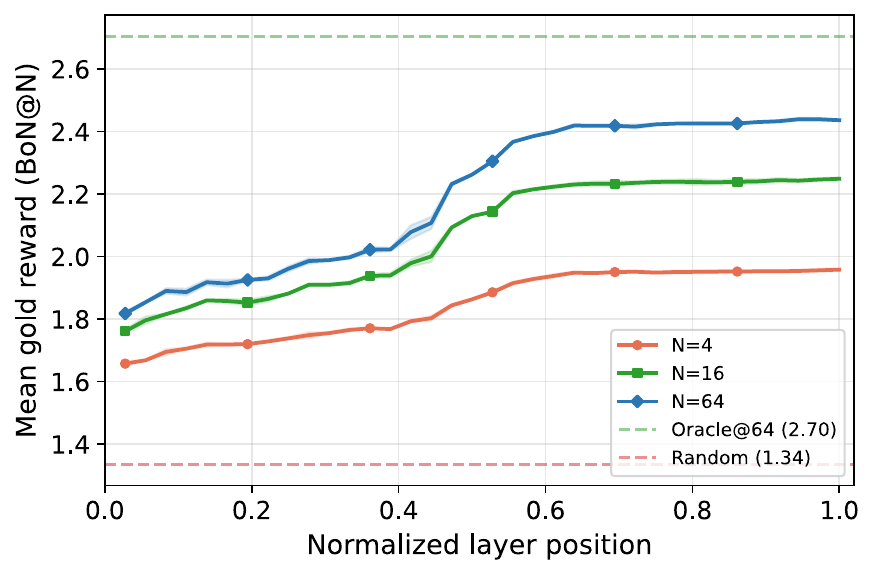}
  \caption{Qwen3-4B}
  \label{fig:n-uf-4b}
\end{subfigure}

\vspace{0.6em}

\begin{subfigure}{0.32\textwidth}
  \centering
  \includegraphics[width=\textwidth]{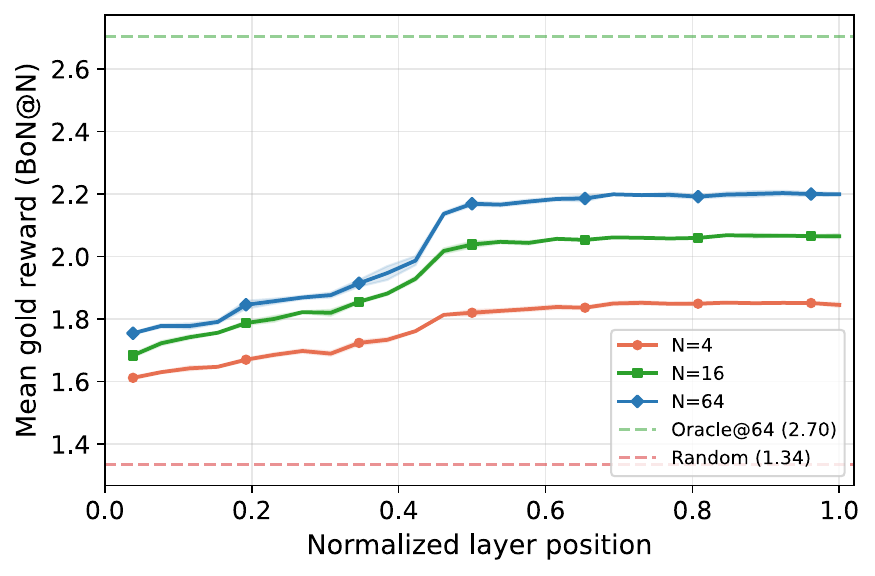}
  \caption{Gemma3-1B}
  \label{fig:n-uf-gemma1b}
\end{subfigure}
\begin{subfigure}{0.32\textwidth}
  \centering
  \includegraphics[width=\textwidth]{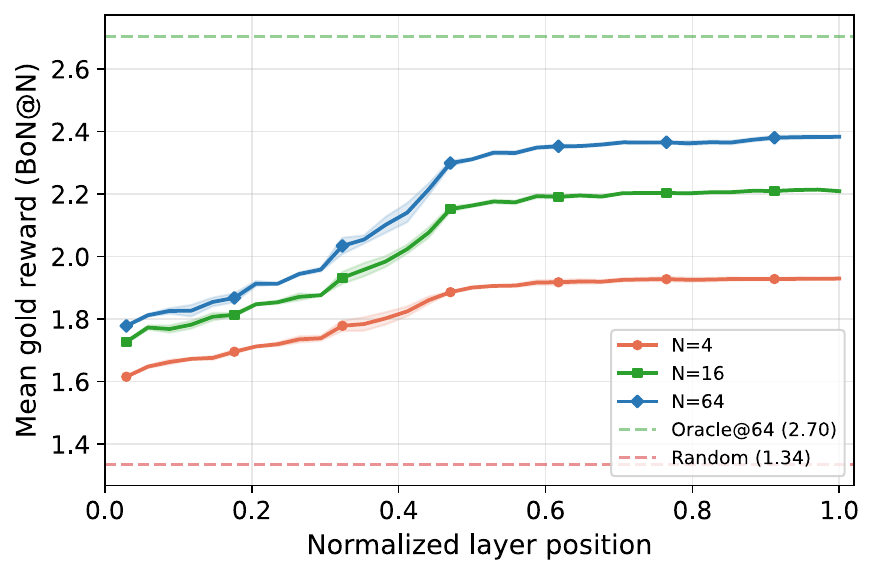}
  \caption{Gemma3-4B}
  \label{fig:n-uf-gemma4b}
\end{subfigure}
\caption{Per-layer BoN at $N \in \{4, 16, 64\}$ on \textbf{UltraFeedback} across model families. Shaded bands: $\pm$1 std across 3 seeds.}
\label{fig:app-n-uf}
\end{figure*}

\begin{figure*}[h]
\centering
\begin{subfigure}{0.32\textwidth}
  \centering
  \includegraphics[width=\textwidth]{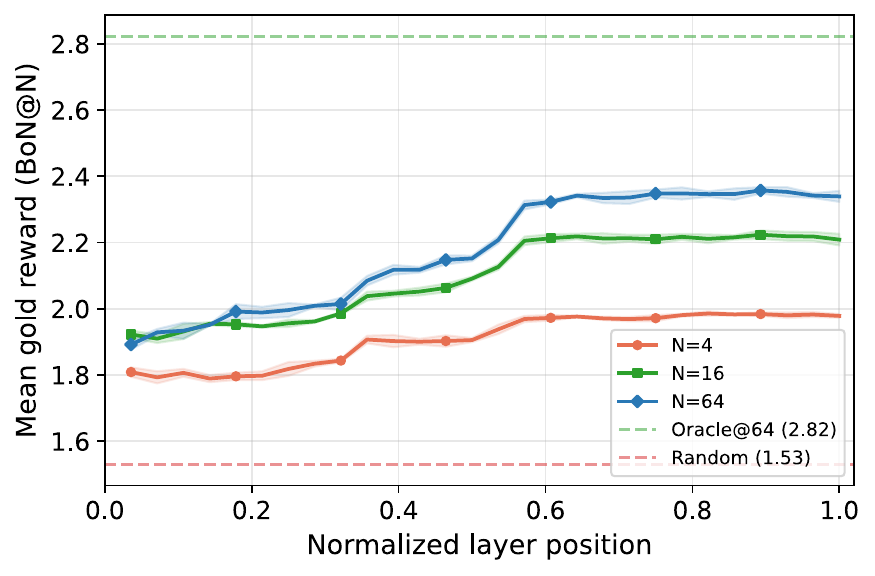}
  \caption{Qwen3-1.7B}
  \label{fig:n-hs2-1.7b}
\end{subfigure}\hfill
\begin{subfigure}{0.32\textwidth}
  \centering
  \includegraphics[width=\textwidth]{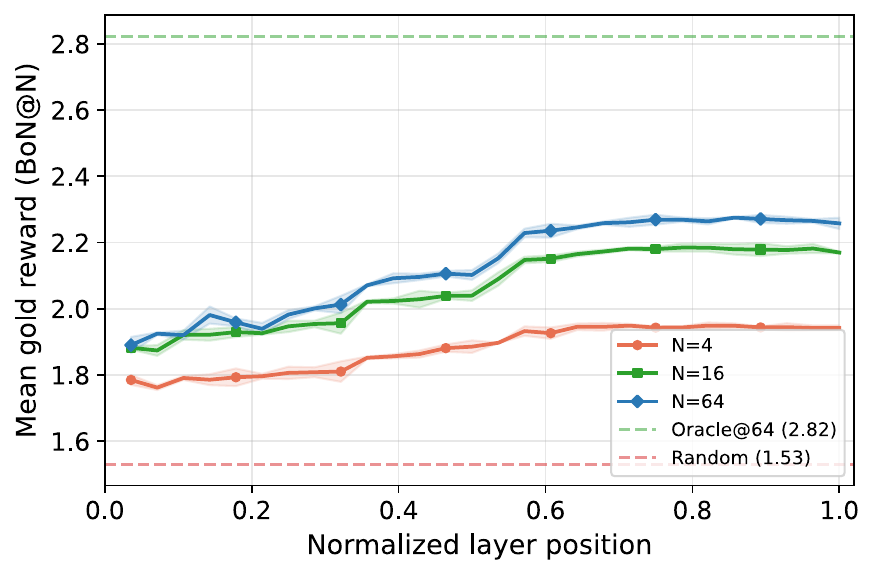}
  \caption{Qwen3-0.6B}
  \label{fig:n-hs2-06b}
\end{subfigure}\hfill
\begin{subfigure}{0.32\textwidth}
  \centering
  \includegraphics[width=\textwidth]{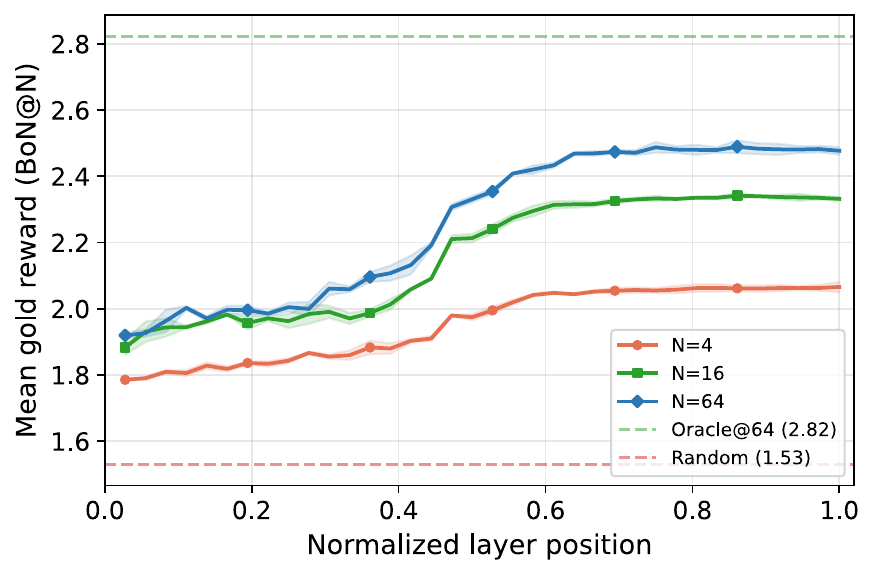}
  \caption{Qwen3-4B}
  \label{fig:n-hs2-4b}
\end{subfigure}

\vspace{0.6em}

\begin{subfigure}{0.32\textwidth}
  \centering
  \includegraphics[width=\textwidth]{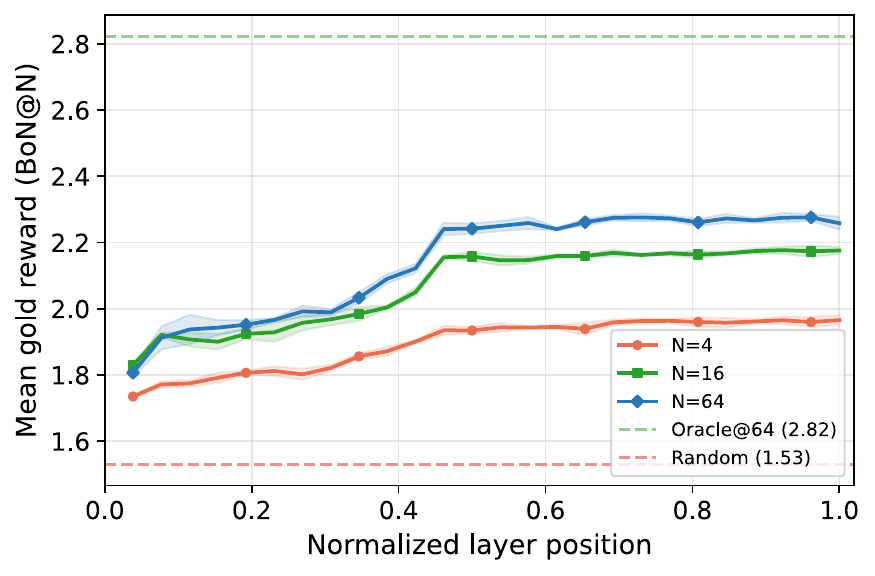}
  \caption{Gemma3-1B}
  \label{fig:n-hs2-gemma1b}
\end{subfigure}
\begin{subfigure}{0.32\textwidth}
  \centering
  \includegraphics[width=\textwidth]{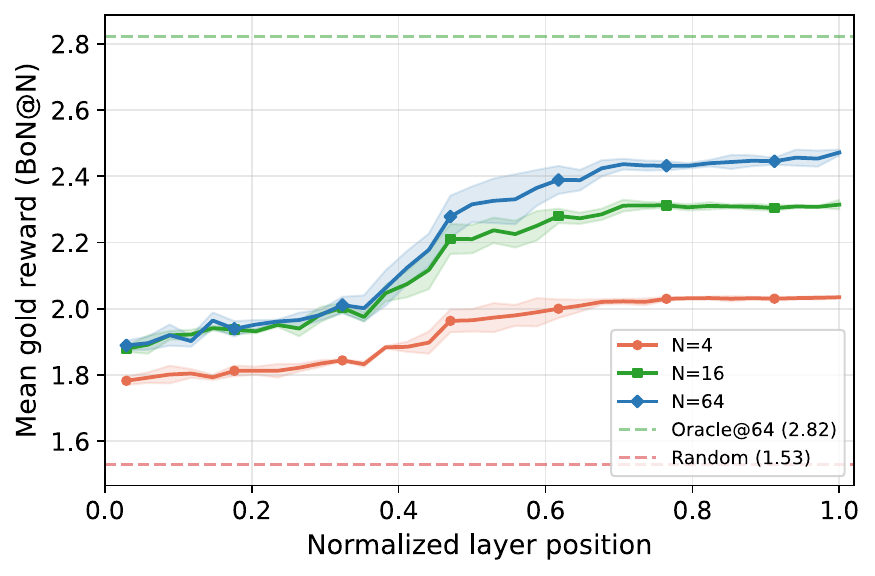}
  \caption{Gemma3-4B}
  \label{fig:n-hs2-gemma4b}
\end{subfigure}
\caption{Per-layer BoN at $N \in \{4, 16, 64\}$ on \textbf{HelpSteer2} across model families. Shaded bands: $\pm$1 std across 3 seeds.}
\label{fig:app-n-hs2}
\end{figure*}

We notice that the per-layer curve shape, RM ranking, and plateau onset are preserved across $N \in \{4, 16, 64\}$ for both datasets, as shown in Figures~\ref{fig:app-n-uf} and \ref{fig:app-n-hs2}, respectively. This is consistent with \textit{(i)}~pipeline structures calibrated at $N{=}12$ transferring to $N{=}64$, which Table~\ref{tab:operating-points} shows directly on the test set; and \textit{(ii)}~deeper layers keeping their advantage when a stage narrows 64 candidates to 16.

\subsubsection{Number of Ranked Responses ($K$)}

\begin{figure}[h]
\centering
\includegraphics[width=.95\columnwidth]{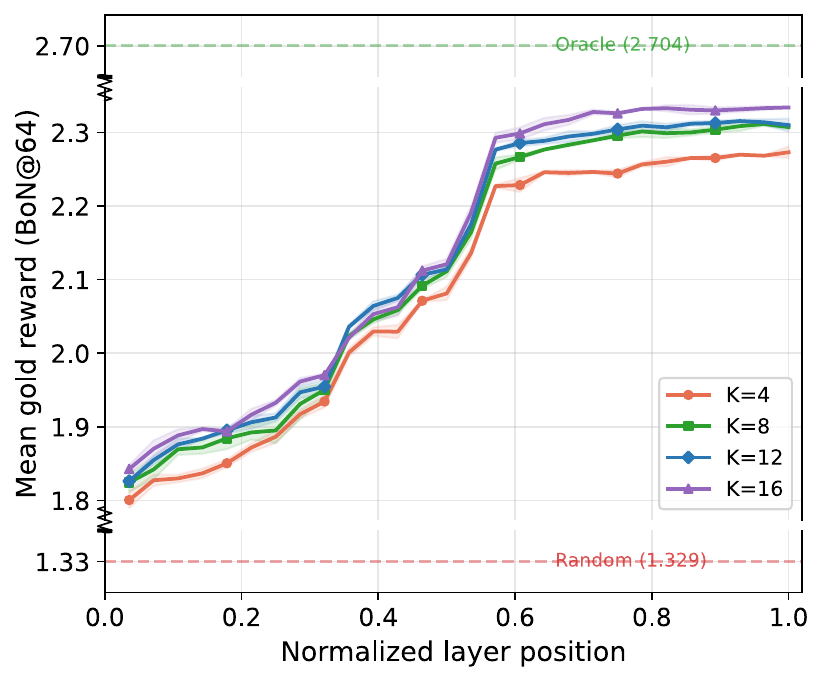}
\vspace{-0.1in}
\caption{Effect of $K$ (number of ranked responses per prompt) on per-layer BoN@64 (Qwen3-1.7B, UltraFeedback). $K{=}4$ underperforms at all layers; $K{=}8$ and $K{=}12$ nearly coincide at the final head, and $K{=}16$ is higher at deep layers.}
\label{fig:numresp}
\vspace{-0.1in}
\end{figure}

Next, we illustrate the performance sensitivity to the number of ranked responses per prompt used to train the exit heads (\S\ref{sec:training}).
As shown on Figure~\ref{fig:numresp}, $K{=}4$ is clearly insufficient. The final-layer BoN drops by 0.04 gold reward compared to $K{=}12$, and early layers suffer more. Going from $K{=}4$ to $K{=}8$ gives the largest gain (+0.034 at the final head), and $K{=}12$ adds only 0.003 over $K{=}8$. On UltraFeedback (27K prompts), $K{=}16$ is better than $K{=}12$ by 0.024 gold reward at the final head; on HelpSteer2 (4.6K prompts), the two are indistinguishable (difference 0.011, against a seed standard deviation of 0.017). We use $K{=}12$ as a robust default.

\subsubsection{Calibration Sample Efficiency}
\label{app:calibration-efficiency}

Subsequently, we test the sensitivity of our results to the selection of the size of calibration set. To this direction, we subsample \{100, 200, 500, 1000, 2000, 2768\} calibration prompts and re-run pipeline optimization at the default $\lambda{=}0.05$ for each size, evaluating blindly on full BoN@64. Results aggregated across 30 runs (2 cells $\times$ 5 RMs $\times$ 3 seeds), 5 random subsamples per size are shown on Table~\ref{tab:calibration-sensitivity}.

\begin{table}[h]
\centering
\small
\begin{adjustbox}{max width=\columnwidth}
\begin{tabular}{lcc}
\toprule
$\bm{N_{\text{cal}}}$ & \textbf{Gap (gold reward)} & \textbf{Gap (\% of range)} \\
\midrule
100 & $-$0.024 & $-$1.72\% \\
200 & $-$0.018 & $-$1.28\% \\
500 & $-$0.006 & $-$0.47\% \\
1000 & $-$0.003 & $-$0.18\% \\
2000 & 0.000 & 0.00\% \\
Full & 0.000 & 0.00\% \\
\bottomrule
\end{tabular}
\end{adjustbox}
\caption{Calibration sample efficiency on UltraFeedback ($\lambda{=}0.05$, 10 configurations $\times$ 3 seeds). Gap = pipeline reward with a subsampled calibration set minus with the full set, in gold reward and in \% of each configuration's oracle$-$random range. With 500 prompts the gap is about 0.5\% of the range, and with $\geq$1000 below 0.25\%. HelpSteer2's calibration set (461 prompts) is too small to subsample at these sizes, but is near the 500 mark.}
\label{tab:calibration-sensitivity}
\vspace{-0.1in}
\end{table}

\subsubsection{Cost Ratio ($\rho$)}
\label{app:cost-sensitivity}

\begin{figure}[h]
\centering
\includegraphics[width=0.95\columnwidth]{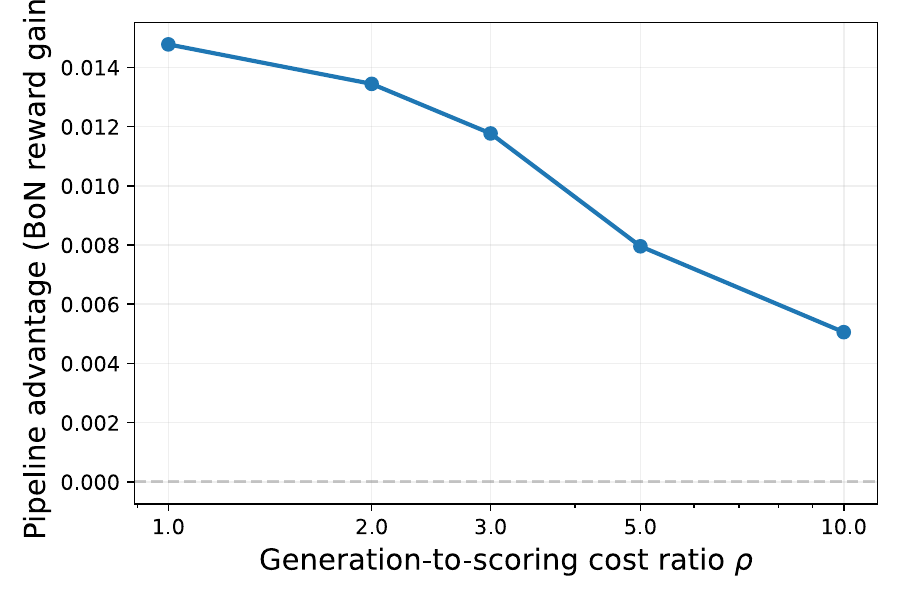}
\caption{Sensitivity of pipeline advantage to generation-to-scoring cost ratio $\rho$. Lower $\rho$ (scoring more expensive) yields larger gains.}
\label{fig:sensitivity}
\end{figure}

The pipeline advantage depends on the generation-to-scoring cost ratio $\rho$, as defined in \S\ref{sec:budget}. In this section, we analyze how the modeling of $\rho$ affects results. Results are depicted in Figure~\ref{fig:sensitivity}.
At $\rho{=}1.0$ (scoring equals generation), gains reach +0.015 gold reward; at $\rho{=}5.0$ (generation dominates), gains are +0.008. The crossover point ($N$ at which pipeline first dominates) shifts earlier at lower $\rho$. The pipeline is most impactful when the RM is large relative to the policy, or when scoring is batched less efficiently than generation. \citet{Gao2023Scaling} found that increasing $N$ can eventually reduce gold reward as selection overoptimizes the proxy RM. At our operating scale ($N \leq 64$), gold reward is still increasing and we observe no saturation.

\subsubsection{Pairwise Threshold Sensitivity ($\tau_{\text{pair}}$)}
\label{app:threshold}

\begin{table}[h]
\centering
\small
\begin{adjustbox}{max width=\columnwidth}
\begin{tabular}{lcccc}
\toprule
$\bm{\tau_{\text{\textbf{pair}}}}$ & \textbf{Training pairs} & \textbf{Final} & \textbf{Layer 14} & \textbf{Layer 4} \\
\midrule
0.5 & 24k & 2.247\small{$\pm$.003} & 2.082\small{$\pm$.003} & 1.885\small{$\pm$.007} \\
1.0 & 21k & 2.243\small{$\pm$.016} & 2.088\small{$\pm$.012} & 1.869\small{$\pm$.006} \\
1.3 & 19k & 2.244\small{$\pm$.005} & 2.096\small{$\pm$.004} & 1.877\small{$\pm$.005} \\
2.0 & 15k & 2.222\small{$\pm$.011} & 2.082\small{$\pm$.003} & 1.881\small{$\pm$.003} \\
\midrule
Multi-response BT & ${\sim}$845k & 2.310\small{$\pm$.009} & 2.114\small{$\pm$.004} & 1.884\small{$\pm$.001} \\
\bottomrule
\end{tabular}
\end{adjustbox}
\caption{Pairwise threshold ablation ($\tau_{\text{pair}}$). Pairwise backbone quality is robust across $\tau_{\text{pair}} \in [0.5, 1.3]$; only $\tau_{\text{pair}}{=}2.0$ degrades slightly at the final layer. Pairwise backbones train on one pair per prompt; the multi-response BT backbone trains on every pair with margin above 0.5 (${\sim}30$ per prompt, margin-weighted). It dominates at deep layers regardless of threshold, which we attribute to its training recipe as a whole.}
\label{tab:threshold}
\vspace{-0.1in}
\end{table}

The pairwise BT baseline uses one chosen/rejected pair per prompt whose gold rewards differ by more than $\tau_{\text{pair}}$ (prompts with a smaller spread are dropped). We use $\tau_{\text{pair}}{=}1.3$, larger than the $\tau{=}0.5$ of our multi-response loss, because every pair contributes equally and must be individually unambiguous. Here, we ablate this threshold at $\tau_{\text{pair}} \in \{0.5, 1.0, 1.3, 2.0\}$ on UltraFeedback / Qwen3-1.7B, train pairwise backbones at each threshold and place identical multi-response BT exit heads on top. Table~\ref{tab:threshold} reports BoN@64 at representative layers.

\FloatBarrier
\subsection{Per-Layer Results under Dataset Shift}
\label{app:ood}

\begin{table}[h]
\centering
\small
\begin{adjustbox}{max width=\columnwidth}
\begin{tabular}{lcc}
\toprule
\textbf{Layer} & \textbf{In-domain (UF)} & \textbf{OOD (HS2)} \\
\midrule
1 & 1.826 $\pm$ .002 & 1.794 $\pm$ .003 \\
10 & 2.036 $\pm$ .003 & 1.909 $\pm$ .019 \\
17 & 2.286 $\pm$ .005 & 2.159 $\pm$ .035 \\
25 & 2.313 $\pm$ .003 & 2.209 $\pm$ .025 \\
final & 2.310 $\pm$ .009 & 2.223 $\pm$ .013 \\
\midrule
\textbf{Layer} & \textbf{In-domain (HS2)} & \textbf{OOD (UF)} \\
\midrule
1 & 1.892 $\pm$ .015 & 1.625 $\pm$ .013 \\
10 & 2.085 $\pm$ .014 & 1.726 $\pm$ .008 \\
17 & 2.322 $\pm$ .007 & 1.979 $\pm$ .009 \\
25 & 2.357 $\pm$ .010 & 1.983 $\pm$ .006 \\
final & 2.339 $\pm$ .017 & 1.980 $\pm$ .011 \\
\bottomrule
\end{tabular}
\end{adjustbox}
\caption{Cross-dataset OOD evaluation (Qwen3-1.7B). Curve shape and plateau depth are preserved; absolute quality degrades.}
\label{tab:ood}
\vspace{-0.1in}
\end{table}

Table~\ref{tab:ood} complements the dataset-shift rows of Table~\ref{tab:shift} with per-layer BoN@64 of the Qwen3-1.7B RM trained on UltraFeedback and evaluated on HelpSteer2, and vice versa.

The curve shape (monotonic climb, plateau at $\sim$60\% depth) is preserved under dataset shift, so the layers that are useful for early exit stay the same, while absolute quality drops. At full depth, comparing each OOD column with the in-domain column of the RM trained on that dataset gives the RM shift loss in column~(a) of Table~\ref{tab:shift}.

\FloatBarrier
\subsection{Pipeline Structures}
\label{app:pipelines}

Tables~\ref{tab:pipe-quality}, \ref{tab:pipe-default} and~\ref{tab:pipe-aggressive} show \emph{what the optimizer selects} at three $\lambda$ values of Table~\ref{tab:operating-points}: quality-focused ($\lambda{=}0.005$), the default ($\lambda{=}0.05$) and a more aggressive $\lambda{=}0.1$, providing reproducible pipeline structures. Table~\ref{tab:operating-points} summarizes savings and quality at every $\lambda$. Pipeline notation: $L_i{\to}K$ means ``layer $i$ retains top-$K$ candidates.'' $K$ values shown are on the calibration set ($N{=}12$); at evaluation ($N{=}64$), $K$ is scaled by $64/12$ (fraction-matching).

\begin{table*}[t]
\centering
\small
\begin{tabular}{llclccr}
\toprule
\textbf{Dataset/Policy} & \textbf{RM} & \textbf{\bm{$S$}} & \textbf{Pipeline (calibration $K$)} & \textbf{Reward} & \textbf{Cost} & \textbf{vs Full} \\
\midrule
\multicolumn{7}{l}{\emph{UltraFeedback / Qwen3-1.7B policy}} \\
 & Qwen3-0.6B & 4 & L10$\to$10, L18$\to$6, L19$\to$2, L24$\to$1 & 2.217 & 0.667 & +0.006 \\
 & Qwen3-1.7B & 4 & L17$\to$8, L19$\to$4, L27$\to$3, final$\to$1 & 2.314 & 0.698 & +0.004 \\
 & Qwen3-4B & 4 & L15$\to$10, L20$\to$6, L31$\to$2, final$\to$1 & 2.440 & 0.718 & +0.004 \\
 & Gemma3-1B & 4 & L12$\to$10, L13$\to$6, L15$\to$3, L24$\to$1 & 2.202 & 0.673 & +0.002 \\
 & Gemma3-4B & 4 & L12$\to$10, L18$\to$6, L20$\to$2, L29$\to$1 & 2.379 & 0.654 & $-$0.004 \\
\midrule
\multicolumn{7}{l}{\emph{HelpSteer2 / Qwen3-1.7B policy}} \\
 & Qwen3-0.6B & 4 & L4$\to$10, L6$\to$8, L16$\to$2, L20$\to$1 & 2.274 & 0.581 & +0.017 \\
 & Qwen3-1.7B & 3 & L11$\to$10, L15$\to$8, L16$\to$1 & 2.328 & 0.602 & $-$0.011 \\
 & Qwen3-4B & 4 & L21$\to$8, L23$\to$3, L26$\to$2, L34$\to$1 & 2.480 & 0.619 & +0.003 \\
 & Gemma3-1B & 4 & L7$\to$8, L10$\to$6, L15$\to$2, L23$\to$1 & 2.276 & 0.483 & +0.017 \\
 & Gemma3-4B & 4 & L12$\to$10, L18$\to$8, L24$\to$2, final$\to$1 & 2.444 & 0.588 & $-$0.028 \\
\midrule
\multicolumn{7}{l}{\emph{UltraFeedback / Gemma3-1B policy}} \\
 & Qwen3-0.6B & 3 & L6$\to$10, L18$\to$3, L25$\to$1 & 1.020 & 0.757 & $-$0.006 \\
 & Qwen3-1.7B & 4 & L17$\to$10, L23$\to$4, L25$\to$2, L27$\to$1 & 1.180 & 0.778 & $-$0.003 \\
 & Qwen3-4B & 4 & L24$\to$4, L27$\to$3, L33$\to$2, final$\to$1 & 1.308 & 0.753 & $-$0.002 \\
 & Gemma3-1B & 4 & L9$\to$10, L17$\to$4, L21$\to$2, final$\to$1 & 0.919 & 0.738 & $-$0.005 \\
 & Gemma3-4B & 4 & L17$\to$8, L19$\to$6, L29$\to$3, L33$\to$1 & 1.227 & 0.726 & +0.002 \\
\midrule
\multicolumn{7}{l}{\emph{HelpSteer2 / Gemma3-1B policy}} \\
 & Qwen3-0.6B & 3 & L16$\to$8, L17$\to$3, L27$\to$1 & 1.713 & 0.607 & +0.004 \\
 & Qwen3-1.7B & 4 & L16$\to$6, L17$\to$3, L22$\to$2, final$\to$1 & 1.882 & 0.627 & +0.006 \\
 & Qwen3-4B & 4 & L20$\to$8, L22$\to$4, L29$\to$3, final$\to$1 & 2.012 & 0.664 & +0.010 \\
 & Gemma3-1B & 4 & L1$\to$10, L13$\to$6, L15$\to$2, final$\to$1 & 1.644 & 0.516 & $-$0.011 \\
 & Gemma3-4B & 4 & L17$\to$6, L21$\to$4, L24$\to$3, final$\to$1 & 1.951 & 0.701 & +0.003 \\
\bottomrule
\end{tabular}
\caption{Optimized pipelines at $\lambda{=}0.005$ (quality-focused). $S$ = number of stages. Cost normalized so full RM on all $N$ = 1.0. Pipeline structures shown for seed 42 as an example; numerical values are means across 3 seeds. 12 of 20 configurations match or exceed the full RM.}
\label{tab:pipe-quality}
\end{table*}

\begin{table*}[t]
\centering
\small
\begin{tabular}{llclccr}
\toprule
\textbf{Dataset/Policy} & \textbf{RM} & \textbf{\bm{$S$}} & \textbf{Pipeline (calibration $K$)} & \textbf{Reward} & \textbf{Cost} & \textbf{vs Full} \\
\midrule
\multicolumn{7}{l}{\emph{UltraFeedback / Qwen3-1.7B policy}} \\
 & Qwen3-0.6B & 4 & L10$\to$10, L14$\to$6, L19$\to$2, L24$\to$1 & 2.206 & 0.569 & $-$0.006 \\
 & Qwen3-1.7B & 4 & L11$\to$10, L16$\to$6, L19$\to$3, final$\to$1 & 2.311 & 0.630 & +0.001 \\
 & Qwen3-4B & 4 & L15$\to$10, L16$\to$8, L23$\to$2, final$\to$1 & 2.437 & 0.648 & +0.000 \\
 & Gemma3-1B & 4 & L10$\to$10, L13$\to$6, L15$\to$3, L24$\to$1 & 2.203 & 0.535 & +0.003 \\
 & Gemma3-4B & 4 & L11$\to$10, L17$\to$4, L20$\to$2, L25$\to$1 & 2.377 & 0.632 & $-$0.005 \\
\midrule
\multicolumn{7}{l}{\emph{HelpSteer2 / Qwen3-1.7B policy}} \\
 & Qwen3-0.6B & 4 & L4$\to$10, L6$\to$8, L16$\to$2, L20$\to$1 & 2.268 & 0.517 & +0.011 \\
 & Qwen3-1.7B & 3 & L11$\to$10, L15$\to$8, L16$\to$1 & 2.320 & 0.553 & $-$0.019 \\
 & Qwen3-4B & 4 & L11$\to$10, L22$\to$4, L25$\to$2, L35$\to$1 & 2.471 & 0.568 & $-$0.006 \\
 & Gemma3-1B & 4 & L7$\to$8, L10$\to$6, L15$\to$2, L23$\to$1 & 2.270 & 0.457 & +0.011 \\
 & Gemma3-4B & 4 & L11$\to$8, L18$\to$6, L23$\to$4, L24$\to$1 & 2.424 & 0.521 & $-$0.048 \\
\midrule
\multicolumn{7}{l}{\emph{UltraFeedback / Gemma3-1B policy}} \\
 & Qwen3-0.6B & 3 & L6$\to$10, L18$\to$3, L25$\to$1 & 1.013 & 0.636 & $-$0.014 \\
 & Qwen3-1.7B & 4 & L17$\to$6, L20$\to$4, L25$\to$2, L26$\to$1 & 1.179 & 0.736 & $-$0.004 \\
 & Qwen3-4B & 4 & L21$\to$6, L24$\to$3, L31$\to$2, final$\to$1 & 1.306 & 0.655 & $-$0.004 \\
 & Gemma3-1B & 4 & L9$\to$10, L17$\to$4, L21$\to$2, final$\to$1 & 0.915 & 0.635 & $-$0.008 \\
 & Gemma3-4B & 4 & L17$\to$6, L24$\to$4, L30$\to$3, L33$\to$1 & 1.227 & 0.677 & +0.001 \\
\midrule
\multicolumn{7}{l}{\emph{HelpSteer2 / Gemma3-1B policy}} \\
 & Qwen3-0.6B & 3 & L16$\to$8, L17$\to$3, L27$\to$1 & 1.716 & 0.587 & +0.007 \\
 & Qwen3-1.7B & 4 & L16$\to$6, L17$\to$3, L22$\to$2, final$\to$1 & 1.882 & 0.627 & +0.006 \\
 & Qwen3-4B & 4 & L20$\to$10, L21$\to$8, L23$\to$2, L27$\to$1 & 2.014 & 0.635 & +0.011 \\
 & Gemma3-1B & 3 & L5$\to$10, L13$\to$6, L14$\to$1 & 1.627 & 0.444 & $-$0.028 \\
 & Gemma3-4B & 4 & L17$\to$6, L21$\to$3, L29$\to$2, final$\to$1 & 1.946 & 0.611 & $-$0.002 \\
\bottomrule
\end{tabular}
\caption{Optimized pipelines at the default $\lambda{=}0.05$. $S$ = number of stages. Cost normalized so full RM on all $N$ = 1.0. Pipeline structures shown for seed 42 as an example; numerical values are means across 3 seeds. Savings range 26--56\%.}
\label{tab:pipe-default}
\end{table*}

\begin{table*}[t]
\centering
\small
\begin{tabular}{llclccr}
\toprule
\textbf{Dataset/Policy} & \textbf{RM} & $\bm{S}$ & \textbf{Pipeline (calibration $K$)} & \textbf{Reward} & \textbf{Cost} & \textbf{vs Full} \\
\midrule
\multicolumn{7}{l}{\emph{UltraFeedback / Qwen3-1.7B policy}} \\
 & Qwen3-0.6B & 4 & L10$\to$10, L14$\to$6, L19$\to$2, L24$\to$1 & 2.209 & 0.517 & $-$0.002 \\
 & Qwen3-1.7B & 4 & L11$\to$10, L16$\to$6, L19$\to$3, final$\to$1 & 2.309 & 0.616 & $-$0.001 \\
 & Qwen3-4B & 4 & L12$\to$10, L16$\to$8, L23$\to$2, final$\to$1 & 2.424 & 0.590 & $-$0.012 \\
 & Gemma3-1B & 4 & L10$\to$10, L13$\to$6, L15$\to$3, L24$\to$1 & 2.203 & 0.535 & +0.003 \\
 & Gemma3-4B & 4 & L11$\to$10, L17$\to$4, L20$\to$2, L25$\to$1 & 2.375 & 0.564 & $-$0.008 \\
\multicolumn{7}{l}{\emph{HelpSteer2 / Qwen3-1.7B policy}} \\
 & Qwen3-0.6B & 4 & L4$\to$10, L6$\to$8, L16$\to$2, L20$\to$1 & 2.268 & 0.517 & +0.011 \\
 & Qwen3-1.7B & 3 & L1$\to$10, L3$\to$8, L16$\to$1 & 2.312 & 0.407 & $-$0.027 \\
 & Qwen3-4B & 4 & L11$\to$10, L19$\to$6, L23$\to$3, L33$\to$1 & 2.472 & 0.562 & $-$0.005 \\
 & Gemma3-1B & 4 & L2$\to$8, L10$\to$6, L12$\to$2, L19$\to$1 & 2.259 & 0.406 & +0.001 \\
 & Gemma3-4B & 4 & L5$\to$10, L11$\to$8, L18$\to$6, L24$\to$1 & 2.421 & 0.513 & $-$0.051 \\
\multicolumn{7}{l}{\emph{UltraFeedback / Gemma3-1B policy}} \\
 & Qwen3-0.6B & 4 & L6$\to$10, L15$\to$8, L18$\to$3, L25$\to$1 & 1.007 & 0.619 & $-$0.019 \\
 & Qwen3-1.7B & 4 & L14$\to$8, L16$\to$6, L19$\to$3, L24$\to$1 & 1.179 & 0.680 & $-$0.004 \\
 & Qwen3-4B & 4 & L19$\to$10, L20$\to$6, L23$\to$2, L28$\to$1 & 1.303 & 0.627 & $-$0.008 \\
 & Gemma3-1B & 4 & L9$\to$10, L17$\to$4, L21$\to$2, final$\to$1 & 0.916 & 0.614 & $-$0.007 \\
 & Gemma3-4B & 4 & L17$\to$8, L18$\to$4, L29$\to$3, L33$\to$1 & 1.221 & 0.624 & $-$0.005 \\
\multicolumn{7}{l}{\emph{HelpSteer2 / Gemma3-1B policy}} \\
 & Qwen3-0.6B & 4 & L13$\to$10, L16$\to$6, L17$\to$4, L18$\to$1 & 1.704 & 0.486 & $-$0.006 \\
 & Qwen3-1.7B & 4 & L16$\to$6, L17$\to$3, L22$\to$2, final$\to$1 & 1.875 & 0.607 & $-$0.001 \\
 & Qwen3-4B & 4 & L17$\to$8, L20$\to$3, L23$\to$2, L27$\to$1 & 2.016 & 0.613 & +0.014 \\
 & Gemma3-1B & 4 & L4$\to$10, L7$\to$8, L12$\to$6, L14$\to$1 & 1.629 & 0.428 & $-$0.026 \\
 & Gemma3-4B & 3 & L16$\to$8, L18$\to$3, L24$\to$1 & 1.940 & 0.581 & $-$0.008 \\
\bottomrule
\end{tabular}
\caption{Optimized pipelines at $\lambda{=}0.1$. $S$ = number of stages. Cost normalized so full RM on all $N$ = 1.0. Pipeline structures shown for seed 42 as an example; numerical values are means across 3 seeds. Savings range 32--59\%.}
\label{tab:pipe-aggressive}
\end{table*}

\clearpage

\begin{table*}[t]
\centering
\small
\setlength{\tabcolsep}{4pt}
\begin{adjustbox}{max width=\textwidth}
\begin{tabular}{llccccccc}
\toprule
\textbf{Exit rule} & \textbf{Shared setting} & \textbf{UF / Qwen} & \textbf{HS2 / Qwen} & \textbf{UF / Gemma} & \textbf{HS2 / Gemma} & \textbf{Mean} & \textbf{Degr.} & \textbf{$<$1\%} \\
\midrule
Best single exit & 68\% depth & 32.1\% & 32.1\% & 32.1\% & 32.1\% & 32.1\% & 0.94\% & 11/20 \\
Confidence gating & entropy, 68\% depth, $t{=}2.3$ & 32.1\% & 32.1\% & 32.1\% & 32.1\% & 32.1\% & 0.94\% & 11/20 \\
PABEE & $p{=}10$ & 12.9\% & 13.2\% & 16.2\% & 10.7\% & 13.2\% & 0.58\% & 13/20 \\
\tool pipeline & $\lambda{=}0.05$ & \textbf{39.7\%} & \textbf{47.7\%} & \textbf{33.2\%} & \textbf{41.9\%} & \textbf{40.6\%} & \textbf{0.34\%} & \textbf{17/20} \\
\bottomrule
\end{tabular}
\end{adjustbox}
\caption{Scoring-FLOP savings of exit rules with one setting shared by all 20 configurations (same RMs and exit heads, 3 seeds). Dataset / policy columns: mean over the five RMs. Each baseline uses its highest-savings setting with mean degradation below 1\%, chosen on test data, so its numbers are an upper bound; the pipeline uses the default $\lambda$. Degr.: mean degradation; $<$1\%: configurations with degradation below 1\%.}
\label{tab:exit-rules}
\end{table*}

\raggedbottom
\makeatletter\global\@colroom\@colht\global\vsize\@colht\makeatother
\section{Qualitative Examples}
\label{app:qualitative}

Here, we present four examples from a two-stage pipeline in which L10 keeps its top 16 of the 64 candidates and L26 picks one of them (UF / Qwen3-1.7B policy, Qwen3-1.7B RM, seed 42). In each, WorldPM scores L10's top-1 pick lower (discrimination fails), the gold-best candidate is among L10's top 16 (filtering works), and L26 selects it. Rank: the candidate's position in L10's ranking of the 64 candidates.

\Needspace{10\baselineskip}
\noindent\textbf{Example 1: Large Gap} (gold reward $\Delta$ = 5.21)

\smallskip\noindent
\colorbox{black!6}{\parbox{0.94\linewidth}{\small\textbf{Prompt:} You will be given a definition of a task first, then some input of the task. In this task, you are given an item and you need to check whether it is edible or not, return 1 if it is edible, else return 2.

cow

Output:}}

\smallskip\noindent
\begin{adjustbox}{max width=\linewidth}
{\small
\begin{tabularx}{\linewidth}{@{}X|X@{}}
\toprule
\textbf{L10 top-1} {\scriptsize (gold: $-3.56$; rank 1/64)} &
\textbf{Gold / L26} {\scriptsize (gold: $1.65$; rank 2/64)} \\
\midrule
2

The cow is not considered edible in the context of the item provided. While the cow's flesh is sometimes consumed, it is generally not classified as an edible item in this specific context. Therefore, the answer is 2. &
1 \\
\bottomrule
\end{tabularx}
}
\end{adjustbox}

\smallskip\noindent
\textit{Analysis:} L10's pick answers 2 (not edible) and justifies it by contradicting itself (``the cow's flesh is sometimes consumed''). L26 picks the correct answer, 1.

\bigskip

\Needspace{10\baselineskip}
\noindent\textbf{Example 2: Medium Gap} (gold reward $\Delta$ = 1.54)

\smallskip\noindent
\colorbox{black!6}{\parbox{0.94\linewidth}{\small\textbf{Prompt:} Provide a brief description of the career path of the individual whose life is described in the paragraph in one sentence. Sarah was a woman who had dedicated her life to advocating for animal rights and worked tirelessly to promote ethical treatment of animals in various organizations.}}

\smallskip\noindent
\begin{adjustbox}{max width=\linewidth}
{\small
\begin{tabularx}{\linewidth}{@{}X|X@{}}
\toprule
\textbf{L10 top-1} {\scriptsize (gold: $0.47$; rank 1/64)} &
\textbf{Gold / L26} {\scriptsize (gold: $2.02$; rank 3/64)} \\
\midrule
Sarah dedicated her life to advocating for animal rights and worked tirelessly to promote ethical treatment of animals in various organizations, shaping the movement through her commitment and efforts. &
Sarah was a dedicated advocate for animal rights, advancing ethical treatment of animals through her work in various organizations throughout her career. \\
\bottomrule
\end{tabularx}
}
\end{adjustbox}

\smallskip\noindent
\textit{Analysis:} L10's pick copies the input sentence nearly word for word and adds an unsupported claim (``shaping the movement''). L26's pick rephrases it as a description of a career, as the prompt asks.

\bigskip

\Needspace{10\baselineskip}
\noindent\textbf{Example 3: Medium Gap} (gold reward $\Delta$ = 1.14)

\smallskip\noindent
\colorbox{black!6}{\parbox{0.94\linewidth}{\small\textbf{Prompt:} You will be given a sentence containing a pronoun/person name and an emotion. From these implicit parameters, the main goal is to find the gender of the person (male / female). \ldots\ [two solved examples] \ldots\ Q: I made Justin feel terrified. A:}}

\smallskip\noindent
\begin{adjustbox}{max width=\linewidth}
{\small
\begin{tabularx}{\linewidth}{@{}X|X@{}}
\toprule
\textbf{L10 top-1} {\scriptsize (gold: $0.92$; rank 1/64)} &
\textbf{Gold / L26} {\scriptsize (gold: $2.06$; rank 4/64)} \\
\midrule
A: male

\textbf{Reason:} ``Justin'' is a common male name, and the context implies the speaker is male (as the sentence is ``I made Justin feel terrified''). The gender is inferred based on the name's typical association with males. &
A: male

\textbf{Reasoning:} The name ``Justin'' is typically associated with a male name. While the emotion ``terrified'' is provided, it does not influence the gender determination in this context. The gender is inferred directly from the name. \\
\bottomrule
\end{tabularx}
}
\end{adjustbox}

\smallskip\noindent
\textit{Analysis:} Both give the same answer. L10's pick adds a claim the sentence does not support (``the speaker is male''); L26's pick bases the answer on the name alone.

\bigskip

\Needspace{10\baselineskip}
\noindent\textbf{Example 4: Small Gap} (gold reward $\Delta$ = 0.28)

\smallskip\noindent
\colorbox{black!6}{\parbox{0.94\linewidth}{\small\textbf{Prompt:} How many wings does a antelope have?}}

\smallskip\noindent
\begin{adjustbox}{max width=\linewidth}
{\small
\begin{tabularx}{\linewidth}{@{}X|X@{}}
\toprule
\textbf{L10 top-1} {\scriptsize (gold: $3.78$; rank 1/64)} &
\textbf{Gold / L26} {\scriptsize (gold: $4.06$; rank 6/64)} \\
\midrule
Antelopes are mammals, and like all mammals, they do not have wings. They are quadrupedal animals with hooves, adapted for running and grazing. While some birds and other animals (like bats) have wings, antelopes are not part of these groups. Therefore, \textbf{an antelope does not have any wings}. &
Antelopes are mammals, and like all mammals, they do not have wings. They are ground-dwelling herbivores with four legs, and their anatomy is adapted for running and grazing. There is no known species of antelope that possesses wings. Therefore, the answer is \textbf{0 wings}. \\
\bottomrule
\end{tabularx}
}
\end{adjustbox}

\smallskip\noindent
\textit{Analysis:} Both are correct and share the same false generalization (bats are mammals with wings); L10's pick names bats in the next sentence, which contradicts it. L26's pick also states the number the question asks for.

\section{Adaptive Exit Rules and Exit Ensembling}
\label{app:negative-results}

In this section, we compare \tool's pipeline with alternatives that use the same trained exit heads: per-prompt adaptive exit rules (\S\ref{app:exit-rules}), a closer look at confidence gating (\S\ref{app:gating}), and combining the scores of several exits (\S\ref{app:ensembling}).

\subsection{Exit-Rule Comparison}
\label{app:exit-rules}

All rules in Table~\ref{tab:exit-rules} use the same trained RMs and exit heads in all 20 configurations and differ only in how they decide where to stop. \emph{Best single exit}: one exit layer scores all candidates. \emph{Confidence gating}: a prompt exits at a fixed layer when a confidence signal of that layer's head passes a threshold, and is scored at full depth otherwise; we use the five signals of \S\ref{app:gating}, with the top-$K$ margin at $K \in \{2, 4, 8, 16\}$ (8 variants), at every layer and 26 thresholds. \emph{PABEE}: patience-based exiting \citep{zhou2020bert} adapted to Best-of-N, which stops at the first layer whose top candidate has been unchanged for $p$ consecutive layers ($p = 2, \ldots, 16$). Each baseline uses one setting for all 20 configurations: the one with the highest mean savings whose degradation, averaged over the 20 configurations, stays below 1\% (3 seeds each). Because the RMs differ in depth (26--36 layers), a setting that names a layer is shared by relative depth: a depth fraction $f$ (from 0.02 to 0.98 in steps of 0.02) maps to layer $\mathrm{round}(f \cdot L)$ of an RM with $L$ exit layers, the same convention as the zero-calibration heuristic. PABEE's patience $p$ counts layers and is shared as is. The pipeline uses the default $\lambda{=}0.05$ without any selection, as in Table~\ref{tab:operating-points}. Choosing the baselines' settings on test data favors them, so their numbers are an upper bound. We do not select them on the calibration set: it has 12 candidates per prompt instead of 64, and the gating threshold and PABEE patience depend on the number of candidates, so settings chosen there do not transfer to the test set.

The best shared gating setting lets every prompt exit at 68\% depth, so it is the best single exit; settings that send some prompts on to full depth save less. The patience rule must score all candidates up to its exit layer and stops too late to save much at this quality level.

\subsection{A Closer Look at Confidence Gating}
\label{app:gating}

We tested five confidence signals: 
\textit{(i)}~score margin ($s_1 - s_N$), 
\textit{(ii)}~entropy of softmax scores, 
\textit{(iii)}~inter-layer agreement (argmax match between adjacent layers), 
\textit{(iv)}~rank correlation (Kendall $\tau$ between layer $L$ and $L{-}k$), and 
\textit{(v)}~top-$K$ margin ($s_1 - s_K$). 
For each, we normalize per layer on the calibration set and sweep a threshold $t$ from 0.5 to 3.0. On UltraFeedback with the Qwen3-1.7B policy and RM (seed 42), the gating setting with the highest savings below 1\% degradation saves 29\% and sends 91\% of prompts out at the same layer (19), scoring the rest at full depth, so it behaves almost like a fixed exit at that layer. The reason is that confidence at an early layer mostly identifies easy prompts, which have high gold reward, but does not predict whether the early layer's top-1 pick matches the final head's pick: the per-prompt cost of exiting early (gold reward gap between the early and the final pick) barely decreases with confidence. This matches the comparison over all 20 configurations in \S\ref{app:exit-rules}, where the best shared gating setting is a fixed exit.

\subsection{Multi-Exit Ensembling}
\label{app:ensembling}

On the same configuration (UltraFeedback, Qwen3-1.7B policy and RM, seed 42), L10 keeps its top 16 of the 64 candidates and we compare strategies for combining the scores of several exits to pick among them: 
\textit{(i)}~a single late layer, 
\textit{(ii)}~the mean of pipeline-stage scores, 
\textit{(iii)}~depth-weighted aggregation,
\textit{(iv)}~the mean of late heads (L20--final), and
\textit{(v)}~Borda count. 
The best-performing variant (mean(L20, L23, L26, final)) achieves 2.3183, matching the full RM (2.3189) to within noise but offering no improvement. Including any earlier layer in the aggregation consistently degrades quality. We further swept the shortlist size $K \in \{4, 8, 12, 16, 24, 32\}$ and found no shortlist size at which ensembling clearly improves on L26 alone. The underlying reason is that adjacent plateau-region heads are too highly correlated ($\tau > 0.83$) for ensembling to shift the argmax.

\section{Data and Model Licensing}
\label{app:licensing}

We summarize the licenses governing the artifacts used in this work.

\paragraph{Models.}
The Qwen3 family models are released under the Apache~2.0 license.
The WorldPM-72B models, which we use as gold reward models, are also Apache~2.0.
The Gemma~3 family models are released under the Google Gemma Terms of Use.

\paragraph{Datasets.}
UltraFeedback is released under the MIT license.
HelpSteer2 is released under CC-BY-4.0.

\end{document}